\documentclass[runningheads]{llncs}
\usepackage{graphicx}
\usepackage[nobiblatex]{xurl}
\usepackage{comment}
\usepackage{rotating}
\usepackage{adjustbox}

\usepackage{amsfonts}

\titlerunning{Improved detection of small objects in road network sequences}

\usepackage[
backend=biber,
style=numeric,
sorting=none
]{biblatex}
\begin{document}
\title{Improved detection of small objects in road network sequences using CNN and super resolution}
%
%
\author{Iv\'an Garc\'ia-Aguilar\inst{1} \and
Rafael Marcos Luque-Baena\inst{1,2} \and
Ezequiel L\'opez-Rubio\inst{1,2}}
\authorrunning{Iv\'an Garc\'ia-Aguilar et al.}
%
\institute{Department of Computer Languages and Computer Science. University of M\'alaga, Bulevar Louis Pasteur, 35, M\'alaga, Spain, 29071 \and
Biomedical Research Institute of M\'alaga (IBIMA), C/ Doctor Miguel D\'iaz Recio, 28, M\'alaga, Spain, 29010\\}
\maketitle              
\begin{abstract}
The vast number of existing IP cameras in current road networks is an opportunity to take advantage of the captured data and analyze the video and detect any significant events. For this purpose, it is necessary to detect moving vehicles, a task that was carried out using classical artificial vision techniques until a few years ago. Nowadays, significant improvements have been obtained by deep learning networks. Still, object detection is considered one of the leading open issues within computer vision. 

The current scenario is constantly evolving, and new models and techniques are appearing trying to improve this field. In particular, new problems and drawbacks appear regarding detecting small objects, which correspond mainly to the vehicles that appear in the road scenes. All this means that new solutions that try to improve the low detection rate of small elements are essential. Among the different emerging research lines, this work focuses on the detection of small objects. In particular, our proposal aims to vehicle detection from images captured by video surveillance cameras.

In this work, we propose a new procedure for detecting small-scale objects by applying super-resolution processes based on detections performed by convolutional neural networks \emph{(CNN)}. The neural network is integrated with processes that are in charge of increasing the resolution of the images to improve the object detection performance. This solution has been tested for a set of traffic images containing elements of different scales to test the efficiency according to the detections obtained by the model, thus demonstrating that our proposal achieves good results in a wide range of situations.

\end{abstract}

\section{Introduction}\label{introduccion}

Currently, object detection is one of the most popular computer vision applications of deep learning. The proliferation of this type of application is caused by the increase of video data obtained from different sources, as well as the improvement in the computational power of the hardware, thereby facilitating the successful accomplishment of this task. The area of road network management can be considered as one of the potential application domains of this type of technology because there are many objects to detect (vehicles), the scenarios are very heterogeneous due to the position of the camera, angle, orientation, and distance to the road, and there is a huge number of already installed traffic cameras with the possibility of capturing relevant information.  

The proposed solution has as its primary objective the detection of reduced size elements (objects) using convolutional neural networks.
Small elements are those which occupy a small region within the entire image. At present, some pre-trained models have significant intrinsic problems that need to be addressed. Object detection models determine the number of features by aggregating information from the raw pixels across the layers of the convolutional network. Most of them reduce the resolution of the images in intermediate layers. This fact causes the loss of the features of small objects, which disappear during the processing carried out by the network, thus avoiding their detection. The low detection rate of this type of element is also caused by the generic background clutter in the images to be inferred, thus making the task of detecting new elements more complex due to many potential object locations. The small objects to be detected have simple shapes that cannot decompose into smaller parts or features. Another point to consider is the quality of the image since there are video surveillance systems with low image quality. Poor quality images negatively affect the performance of object detection methods, because these types of sequences, especially in the case of traffic videos, are dense, with small vehicles and partially occluded. On the other hand, some objects are similar in feature, shape, color, or pattern. Therefore, existing pre-trained high-quality models cannot distinguish among them accurately.

\thinspace
The small object detection problem is relevant in many areas. Direct applications include the following: the enhancement in the classification and detection of elements through images captured by satellites, the improvement in video surveillance through the treatment of images provided by security cameras established at high points, and the increase in the detection of pedestrians or traffic. This article focuses on the last of these direct applications. Thanks to this new solution, it will be possible to detect a larger number of elements and improve the confidence of each detection without the need to retrain the model so that this solution may be advantageous for automated traffic control systems.

\thinspace
Currently, there are some models aimed at the detection of these elements through two workflows. The first of them follows the usual flow in which a series of candidate regions are generated to perform the classification of each proposed area subsequently. The second method establishes the detection of objects as a regression or classification problem to adopt a framework to achieve final results. Among the methods based on regional proposals, we find mainly R-CNN, Faster R-CNN, and Mask R-CNN, while in regression methods, we can mention SSD and YOLO \cite{modelos}. As an example, we find the model called \emph{EfficientDet}, which has an average detection rate of 51\% for medium-sized elements. In contrast, smaller elements are detected in only 12\% of the cases. It should also be noted that there are no datasets devoted explicitly to the detection of small objects, which is why most of the attained detections correspond to larger elements. When applying existing models such as \emph{Faster-RCNN} it is found that it misses several small objects in part due to the size of the anchor boxes of the elements detected in the image.

\thinspace
The proposal put forward in this article is based on the design, implementation, and subsequent testing of a technique based on convolutional neural networks, capable of detecting small-scale elements and improving the class inference, without the need to retrain the pre-trained model. To achieve this goal, the hyper-parameters of the network have been modified to reduce false negatives. Subsequently, super-resolution processes have been applied to the input image to generate a series of images centered on each of the detected elements to increase its resolution. Then the model yields an enhanced inference about the object class. Using this method, implemented in conjunction with image preprocessing through a denoising filter, small object detection and class inference are significantly improved. 

\begin{figure}[ht!]
\centering
\includegraphics[width=1\linewidth]{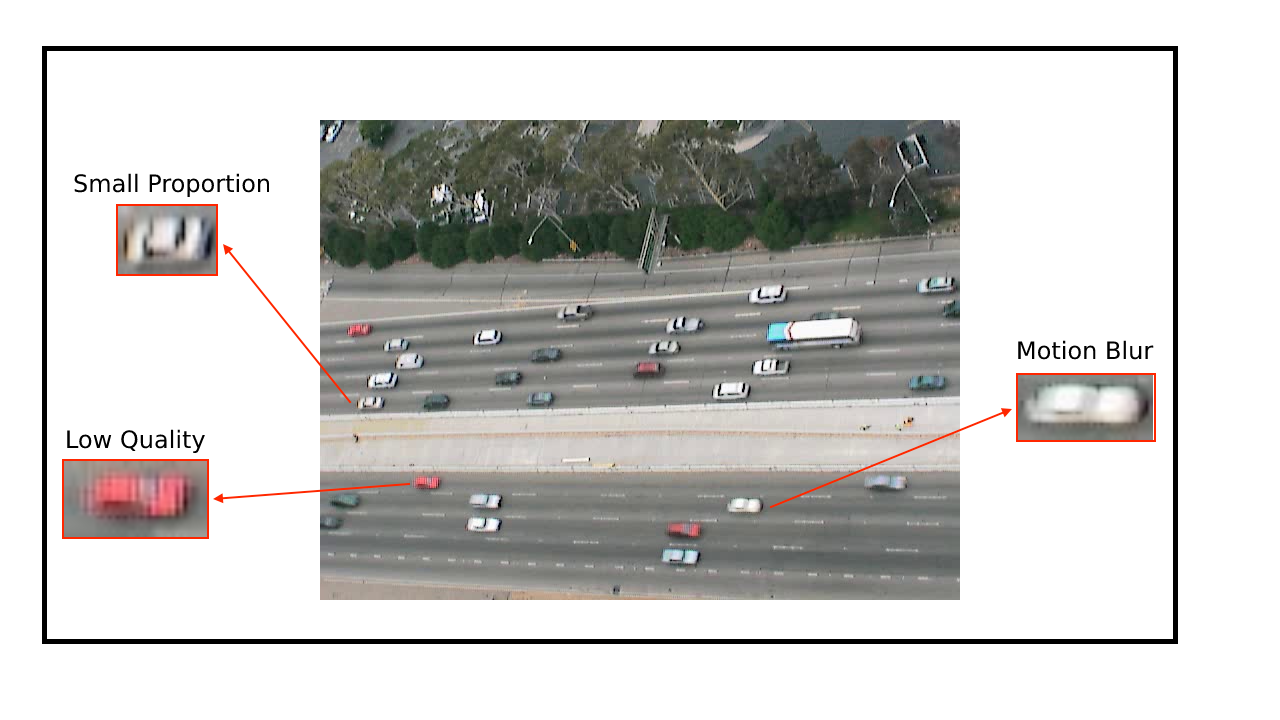}
\caption{Main challenges to be solved}
\label{fig:f0}
\end{figure}

Additionally, it has been necessary to create a new dataset to perform the relevant tests to evaluate the qualitative improvement attained by the newly developed technique. Most existing datasets are not adequate, such as \emph{KITTI} \cite{Geiger2013IJRR}, which is composed of a series of images collected by cameras located on mobile vehicles. This set of images contains several vehicles. However, the size of these images is large, and it should be noted that not all images include vehicles, and in other cases, the number of vehicles is small. It should be noted that nowadays, traffic-oriented video surveillance systems produce sharp and clear images. With this in mind, it can therefore be established that there are no appropriate public datasets on which to perform tests, so a new dataset of vehicle detection data has been developed from a series of traffic videos captured by cameras and surveillance systems intended for that purpose. Three test sets filmed in different locations and with a series of challenges such as image quality, light interference, motion blur, and light interference, among others, have been compiled. This dataset consists of 476 images with a total of 14557 detections. These images contain a set of challenges to be solved, as shown in Figure \ref{fig:f0}.  

\thinspace
Under the premises described above, it can be concluded that there are significant shortcomings in the state of the art methods for the detection of small-scale elements, given the low detection rate of current models and the few performance-enhancing procedures.

\thinspace
The rest of this article is organized as follows. Section 2 sets out the related work. Throughout Section 3, the improvements developed are detailed, explaining in depth the implemented workflow. Section 4 includes the performed tests along with their respective results. Finally, in Section 5, the conclusions and the future works to be developed according to the proposed solution are outlined.

\section{Related work}

Given the context of our proposal, computationally intensive image processing is required. Papers such as (\cite{girshick2015fast}; \cite{8113363}) have identified new methods that significantly accelerate the detection of visual elements, thus allowing high-speed image inference.

Currently, there are pre-trained models such as, for example, \emph{YOLO} \cite{redmon2016look}. This model divides the image into a grid to subsequently predict the class labels for each bounding box. Another frequently used model is \emph{CenterNet}. This model detects each of the objects as a triplet instead of a pair, thus improving the accuracy in detecting and inferring each element's class.

According to previous studies, it has been established that methods based on regional proposals obtain a higher hit rate when detecting elements in an image. On the other hand, it should be noted that there are some detectors based on \emph{RCNNs} such as (\cite{modelo1};  \cite{modelo2}). However, the detection of small-scale elements is referenced without going into great detail. Other works such as (\cite{a1}; \cite{a2}) study the modification of the model architecture to detect small size elements such as faces and pedestrians, respectively. Precisely in detecting small elements such as pedestrians, there are some advances and interesting developments. For example, \cite{numero47} proposes a pedestrian detector that operates with multilayer neural representations to predict pedestrian locations accurately. Also, \cite{numero48} presents a new knowledge distillation framework to learn a pedestrian detector that reduces the computational cost associated with detections while maintaining high accuracy. In the field of vehicle detection, we can highlight works such as the one proposed by \cite{vehiculo1} that recognizes the direction of the vehicle in the input image. This method may be suitable to be able to classify the vehicle model based on vehicle appearance. Nevertheless, as the distance from the vehicle gets larger, the image quality deteriorates, thereby resulting in a worse accuracy in recognizing small elements.

On the other hand, image super-resolution processes establish a series of techniques to increase the resolution of an input image. There are several advances and developments for super-resolution processes applied to a single image. For example, \cite{articlesr1} proposed a network based on super-resolution processes known as \emph{SRCNN}. This network establishes an end-to-end mapping between images given as input with low resolution and those processed by the model. Furthermore, \cite{articlesr2} proposes a network known as \emph{VDSR} which stacks a series of convolutional layers with residual character learning. Also, \cite{articlesr3} proposed a network denoted as \emph{EDSR} which achieves high performance for \emph{SR} processes by eliminating batch normalization layers, resulting in the model known as \emph{SRResNet}.

Although these processes remain at a primitive stage and are complex, there are significant advances, mainly focused on the creation of algorithms that perform these functions effectively. One of the main approaches in this field is presented by \cite{7115171}, as they proposed super-resolution algorithms based on deep learning. One of the most notable applications in the area of super-resolution corresponds to the processing of satellite images, wherein  \cite{bosch2017superresolution}  establishes a quantitative analysis to determine the success of improvements in the resolution of these images through the modification of layers in the detection model. Finally, it is worth mentioning articles that are related to our work. First of all, \cite{Cao2016VehicleDF} uses high-resolution aerial imagery to improve vehicle detection using a simple linear model. In that article, a significant improvement of the results is obtained thanks to super-resolution processes. Works such as \cite{articlepapersr} advocate the generation of preprocessed low-resolution images to increase the coverage in the detections inferred by the model.

Furthermore, \cite{sr1} proposes a method based on super-resolution using deep neural networks based on \emph{GANs} to reconstruct a low-resolution image and obtain a new image that improves license plate recognition. Throughout this paper, a super-resolution model is developed which makes use of deep residual blocks. They use adversarial loss allowing the discriminator to distinguish between the input image and the image processed with the super-resolution model. On the other hand, the perceptual loss is used to help the generator better reconstruct the high-resolution image. The license plate recognition improves from 38.45\% to 73.65\% by the application of this method.

In addition to this, \cite{improves} proposes super-resolution processes based on a series of regions detected by an \emph{RPN}. These are applied to improve the detection and classification of visual elements.

Although these approaches are related to the one presented in this paper, they fail to propose a scheme based on current convolutional neural network models. Thanks to the super-resolution processes as well as the pre-processing stage, our proposal finds out new visual elements. In this paper, we hypothesize an improvement in the detection of small-scale objects using super-resolution (\emph{SR}) techniques for the enhancement in the inference provided by models such as \emph{CenterNet HourGlass}, specifically for sequences obtained by video surveillance cameras set at high points.

\section{Methodology\label{sec:Methodology}}

\begin{figure}[ht!]
\centering
\includegraphics[width=1\linewidth]{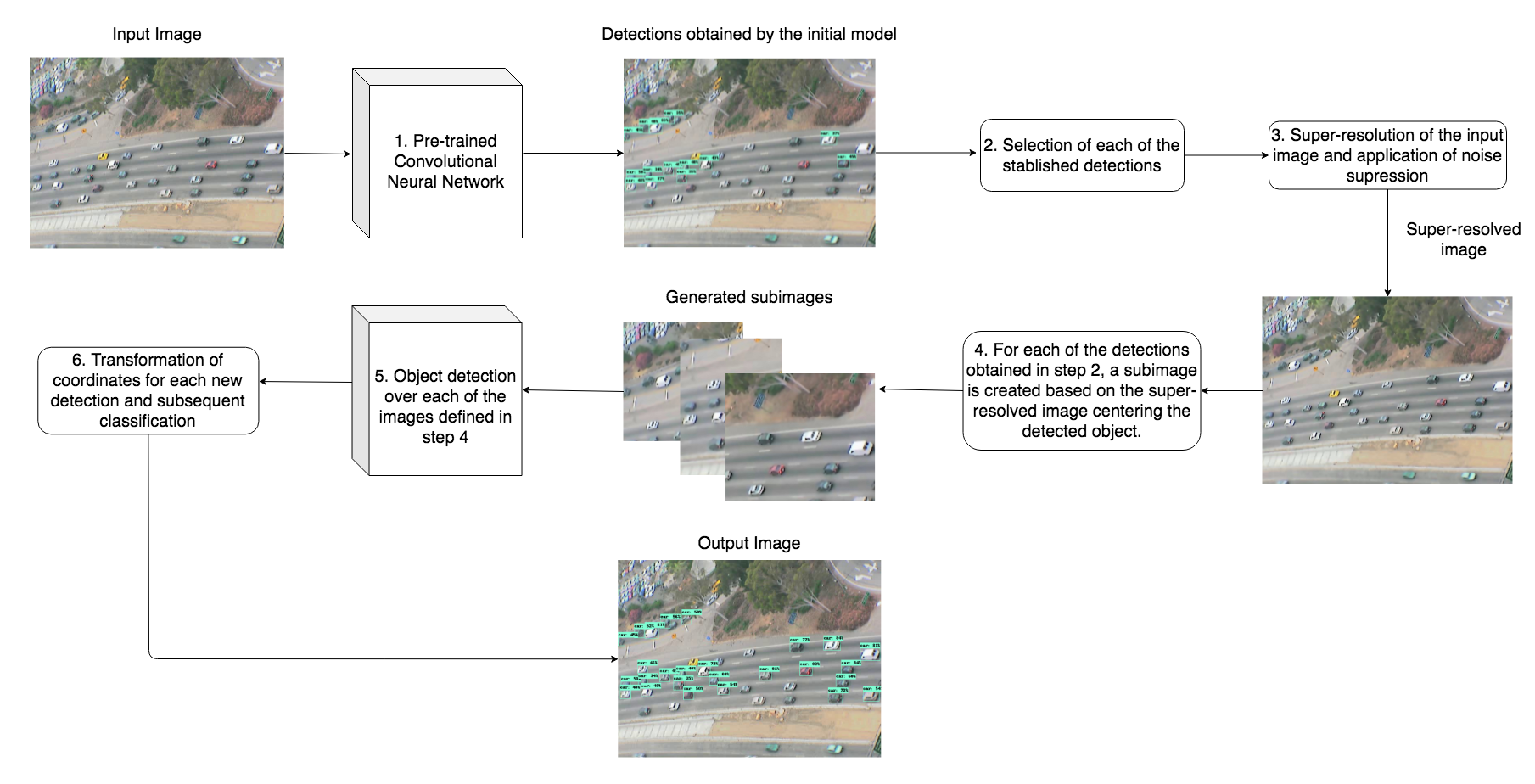}
\caption{Workflow of the proposed technique}
\label{fig:fworkflow}
\end{figure}

Next, our small object detection system is presented. Figure \ref{fig:fworkflow} shows the global workflow of our proposed framework.

\begin{figure}[ht!]
\centering
\includegraphics[width=1\linewidth]{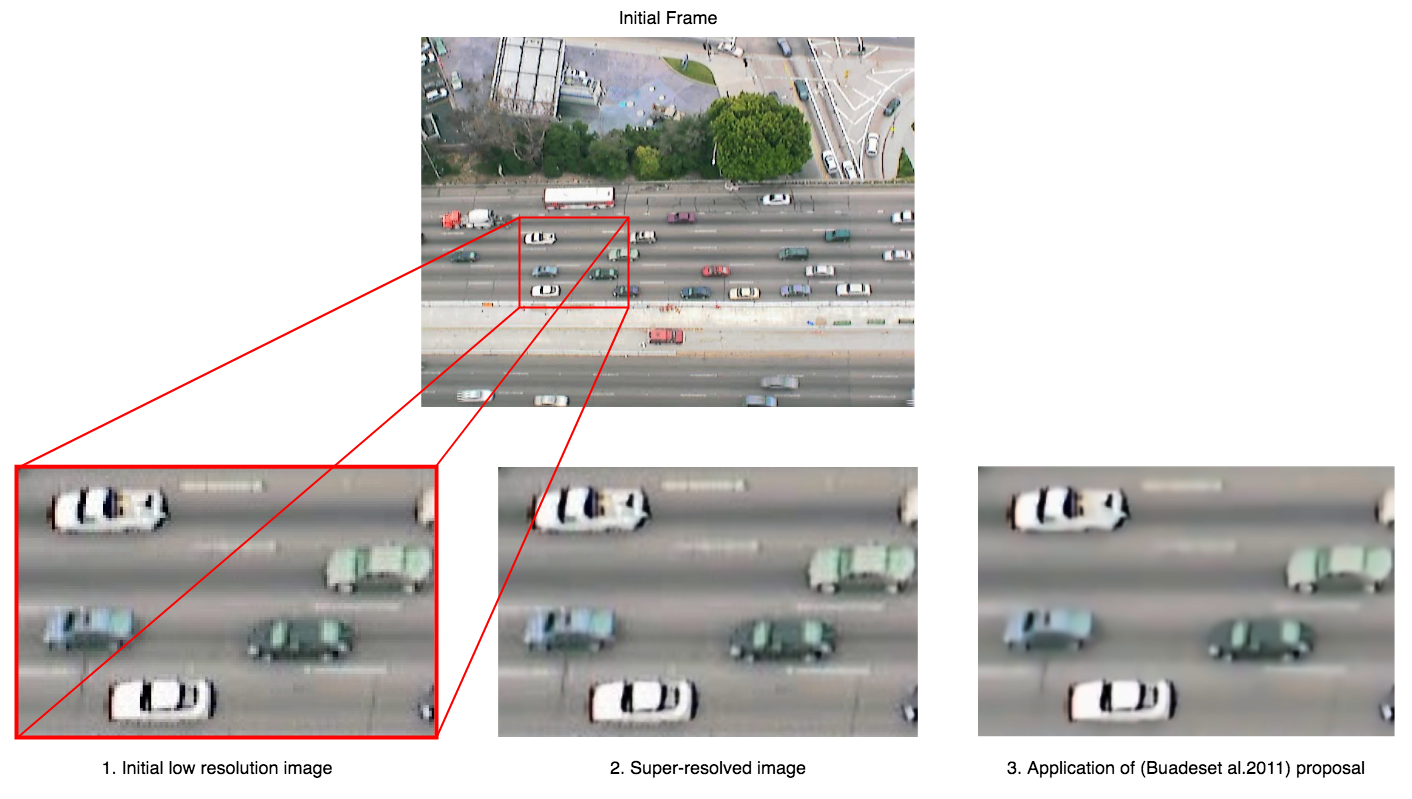}
\caption{Frame processing}
\label{fig:fComplete}
\end{figure}

In what follows, the methodology of our proposal is explained in detail. As set out in point 1 of the workflow, let us consider a deep learning neural network for object detection which
takes an image $\mathbf{X}$ as input and returns a set of detections
$S$ as output:

\begin{equation}
S=\mathcal{F}\left(\mathbf{X}\right)
\end{equation}

\begin{equation}
S=\left\{ \left(a_{i},b_{i},c_{i},d_{i},q_{i},r_{i}\right)\;|\;i\in\left\{ 1,...,N\right\} \right\} 
\end{equation}

where $N$ is the number of detections, $ \left(a_{i},b_{i}\right)\in\mathbb{\mathbb{R}}^{2} $
are the coordinates of the upper left corner of the i-th detection
within the image $\mathbf{X}$, $\left(c_{i},d_{i}\right)\in\mathbb{\mathbb{\mathbb{R}}}^{2}$
are the coordinates of the lower right corner of the $i$-th detection
within $\mathbf{X}$, $q_{i}$ is the class label of the detection,
and $r_{i}\in\mathbb{R}$ is the class score of the detection. The
higher $r_{i}$, the more confidence that an object of class $q_{i}$
is actually there. It is assumed that the origin of the coordinate
system of all images is at the center of the image. Next, step two is performed based on the selection of each of the initially established detections. 
Given a low-resolution input image $\mathbf{X}_{LR}$ as shown, for example, in figure \ref{fig:fComplete}, our first step is to process it with the object detection network to yield a set
of tentative detections $S_{LR}$:

\begin{equation}
S_{LR}=\mathcal{F}\left(\mathbf{X}_{LR}\right)
\end{equation}

Subsequently, as detailed in the third step of the workflow, a new image is generated by applying super-resolution and denoising processes to the initial image given as input.

A super-resolution deep network is used to obtain a high-resolution
version $\mathbf{X}_{HR}$ with zoom factor $Z$ of the low-resolution
input image $\mathbf{X}_{LR}$ as shown in point 2 of the figure \ref{fig:fComplete}. To implement this super-resolution process, the OpenCV library has been used. There are some pre-trained models for the execution of processes that increase the initial resolution of an image. The considered models are the following:

\begin{itemize}
    \item Fsrcnn: Fast Super-Resolution Convolutional Neural Network \cite{fsrcnn}.
    \item Edsr: Enhanced Deep Residual Networks Single Image Super-Resolution \cite{articlesr3}.
    \item Espcn: Efficient Sub-Pixel Convolutional Neural Network \cite{espcn}.
    \item Lapsrn: The Laplacian Pyramid Super-Resolution Network \cite{lapsrn}.
\end{itemize}

According to the final image quality obtained by each of these models as well as the time required to carry out the process, the \emph{Fsrcnn} model has been selected because it is one of the models that process the images with the best speed and quality.

Subsequently, to improve the detections by the network, noise elimination is performed. This process is performed through the use of the non-local denoising algorithm\footnote{\url{http://www.ipol.im/pub/algo/bcm\_non\_local\_means\_denoising/}} making use of various computational optimizations \cite{buades_2011}. This method is applied when Gaussian white noise is expected (Point 3 of the Figure \ref{fig:fComplete}). After an empirical study, it was concluded that this type of processing improves the number of detected elements.

Then, for each detection in $S_{LR}$,
a subimage $\mathbf{X}_{i}$ with the same size as $\mathbf{X}_{LR}$
is extracted from $\mathbf{X}_{HR}$. The subimage $\mathbf{X}_{i}$
is centered at the center of the detection:

\begin{equation}
\mathbf{y}_{i}=\left(\frac{a_{i}+c_{i}}{2},\frac{b_{i}+d_{i}}{2}\right)
\end{equation}

\vspace{1em}

\begin{equation}
\hat{\mathbf{y}}_{i}=Z\mathbf{y}_{i}
\end{equation}
where $\mathbf{y}_{i}$ is the center of $\mathbf{X}_{i}$ expressed
in coordinates of $\mathbf{X}_{LR}$, while $\hat{\mathbf{y}}_{i}$
is the center of $\mathbf{X}_{i}$ expressed in coordinates of $\mathbf{X}_{HR}$.
The object detection network is applied to $\mathbf{X}_{i}$ in order
to yield a new list of detections:

\begin{equation}
S_{i}=\mathcal{F}\left(\mathbf{X}_{i}\right)
\end{equation}

As indicated in step 5 of the workflow, each of the generated images is again given to the object detection model to improve the inference or detect elements not initially spotted.

The detections of $S_{i}$ are expressed in coordinates of $\mathbf{X}_{i}$,
so that they must be translated to coordinates of $\mathbf{X}_{LR}$.
The equation to translate a point $\tilde{\mathbf{h}}$ in coordinates
of $\mathbf{X}_{i}$ to coordinates $\mathbf{h}$ of $\mathbf{X}_{LR}$
is:

\begin{equation}
\mathbf{h}=\mathbf{y}_{i}+\frac{1}{Z}\tilde{\mathbf{h}}
\end{equation}

\begin{figure}[ht!]
\centering
\includegraphics[width=0.8\linewidth]{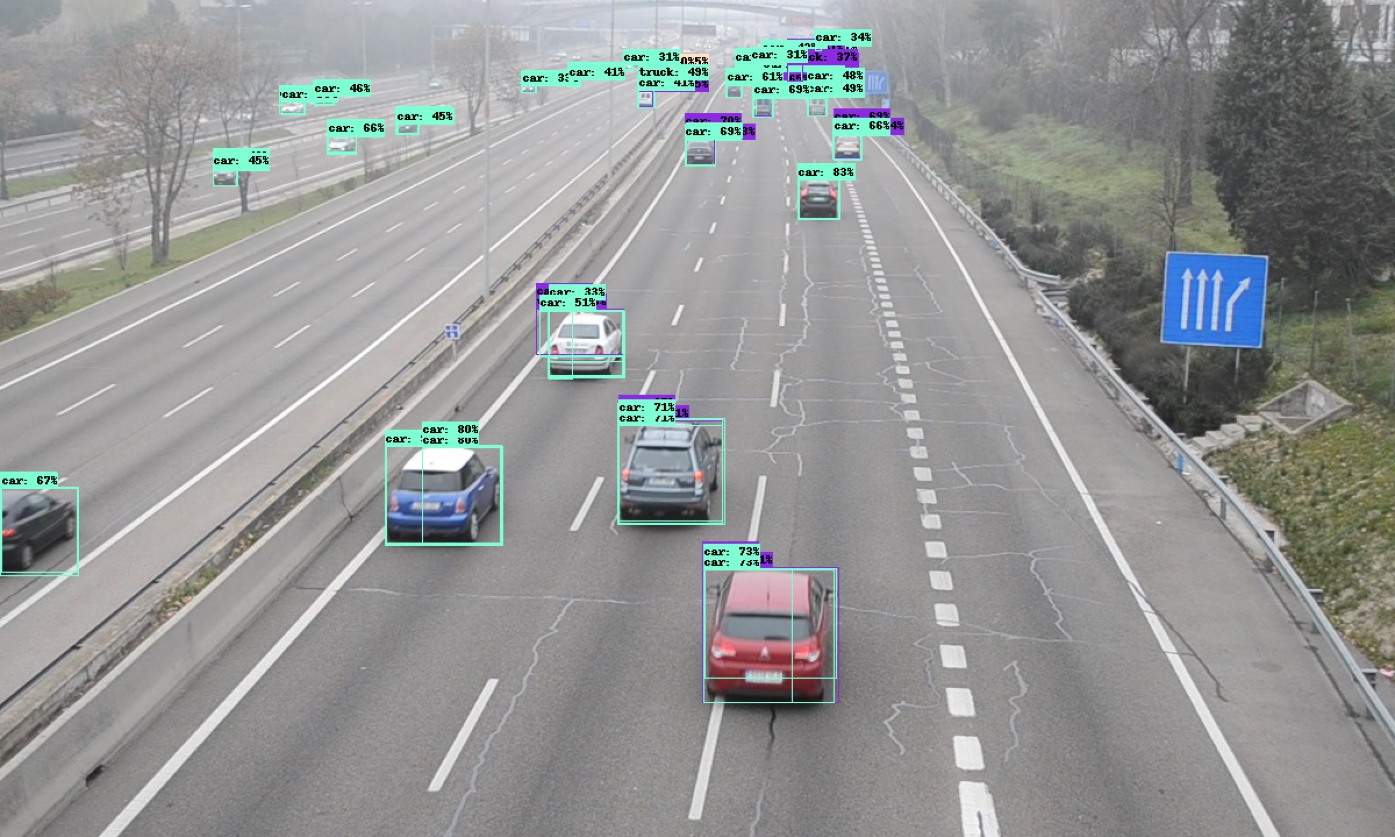}
\caption{Video frame before filtering matching detections}
\label{fig:f3}
\end{figure}

As shown in Figure \ref{fig:f3}, for the same object there will be multiple possible detections before performing the intersection over union (IOU) operation.

After that, all the detections coming from the detection set $S_{i}$
are processed to estimate whether they correspond to new objects,
or they match some detection already present in $S_{LR}$. To this
end, the intersection over union measure is computed for each pair of detections $D_{j}$ and $D_{k}$:

\begin{equation}
IOU=\frac{Area\left(D_{j}\cap D_{k}\right)}{Area\left(D_{j}\cup D_{k}\right)}
\end{equation}

\vspace{1em}

\begin{figure}[ht!]
\centering
\includegraphics[width=0.8\linewidth]{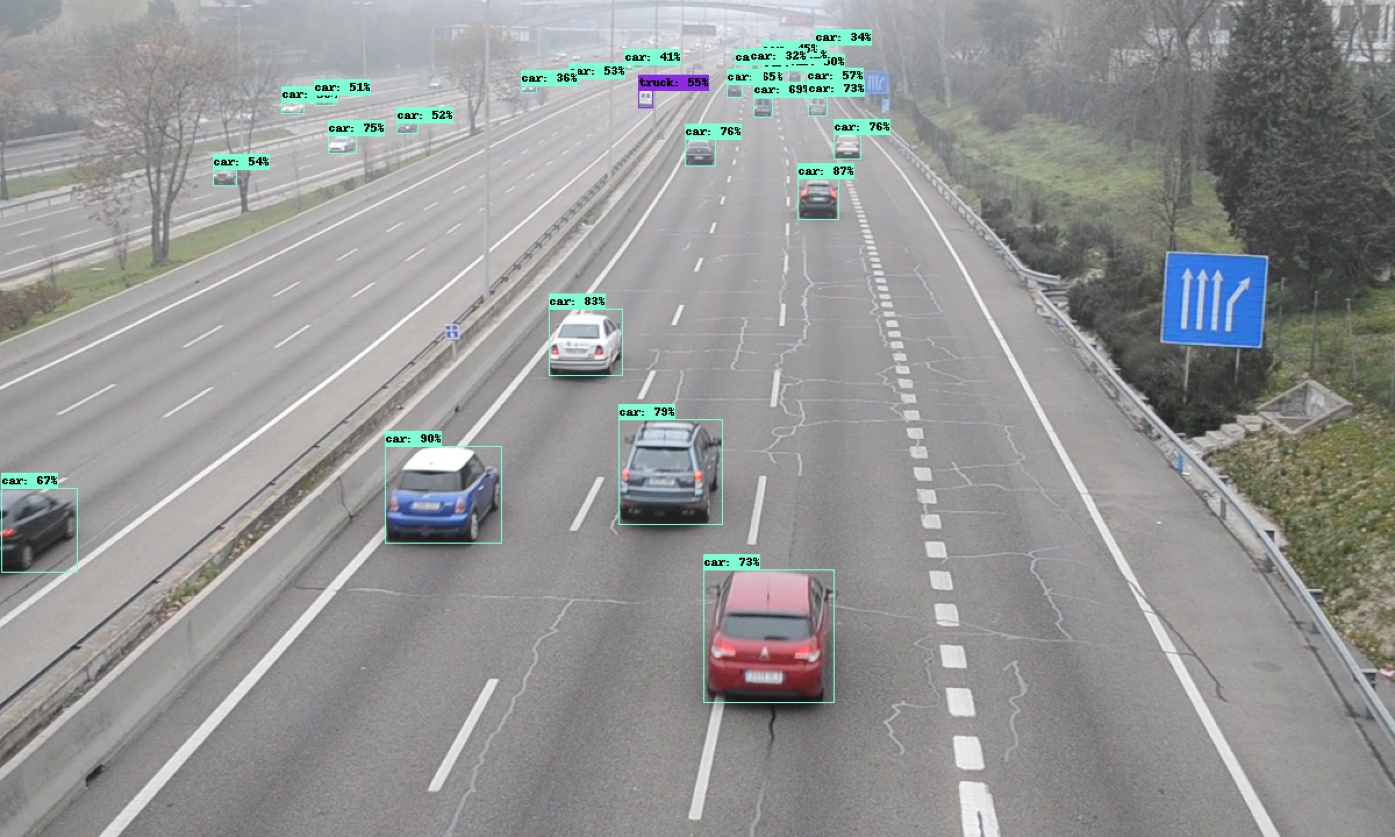}
\caption{Final result after filtering matching detections}
\label{fig:f4}
\end{figure}

The detections $D_{j}$ and $D_{k}$ are judged to correspond to the same object whenever $IOU>\theta$, where $\theta$ is a tunable parameter.
The final set of detections is comprised of those detections that
persist after filtering the matching detections. This point finally corresponds to step 6 of the workflow. At the end of this process, an image will be obtained with a higher number of detections and with an improvement in the class inference of each element, see Figure \ref{fig:f4}.

\section{Experimental results}\label{results}

In order to carry out the tests on the data set designed for the experiments, it is necessary to establish a series of parameters such as the maximum number of detected objects, the minimum inference to take into account an object as well as the threshold $\theta$ for the intersection over union index to eliminate simultaneous detections for the same object. These parameters are established in Table \ref{tabla1} and selected according to the best results obtained after an empirical study.

\vspace{1em}

\begin{table}[!ht]
\begin{center}
    \begin{tabular}{| c | c |}
\hline
Parameter & Value \\ \hline
Maximum number of detections & 100 \\ \hline
Minimum percentage of inference & 0.3 \\ \hline
IOU threshold $\theta$ & 0.1 \\ \hline
\end{tabular}
\caption{Selected values of the hyperparameters.}
\label{tabla1}
\end{center}
\end{table}

\vspace{1em}

Regarding the dataset used for the tests, as stated in the Introduction (Section \ref{introduccion}), three videos from video surveillance systems were manually labeled, all of which pointed to roads with a large number of vehicles, video one is named as \emph{sb-camera1-0820am-0835am}, video two as  \emph{sb-camera3-0820am-0835am} and video three as \emph{sb-camera4-0820am-0835am}. As can be seen in the name of each of the videos, they are formed by the number of the camera referred to as well as the time that the images were acquired. This dataset consists of 476 images with a total of 14557 detections. Four specific categories related to the detection of vehicles on roads have been selected. Namely, the labels \emph{car}, \emph{truck}, \emph{motorcycle}, and \emph{bus} have been chosen to analyze the results. The dataset is composed of a series of images with specific interesting characteristics for the study of the obtained results, mainly highlighting the following aspects:

\begin{itemize}
    \item The vehicles set in each frame occupy a small area of the image, thus qualifying them as small size elements. In datasets such as \emph{KITTI}, we have a large number of vehicles. However, the size of the vehicles is much larger compared to the dataset manually developed for this article.
    \item There is an imbalance of classes. There are more abundant categories of elements such as cars versus motorcycles which appear in a limited number of frames.
\end{itemize}

\begin{figure}[!ht]
   \begin{minipage}{0.495\textwidth}
     \centering
     \includegraphics[width=1\linewidth]{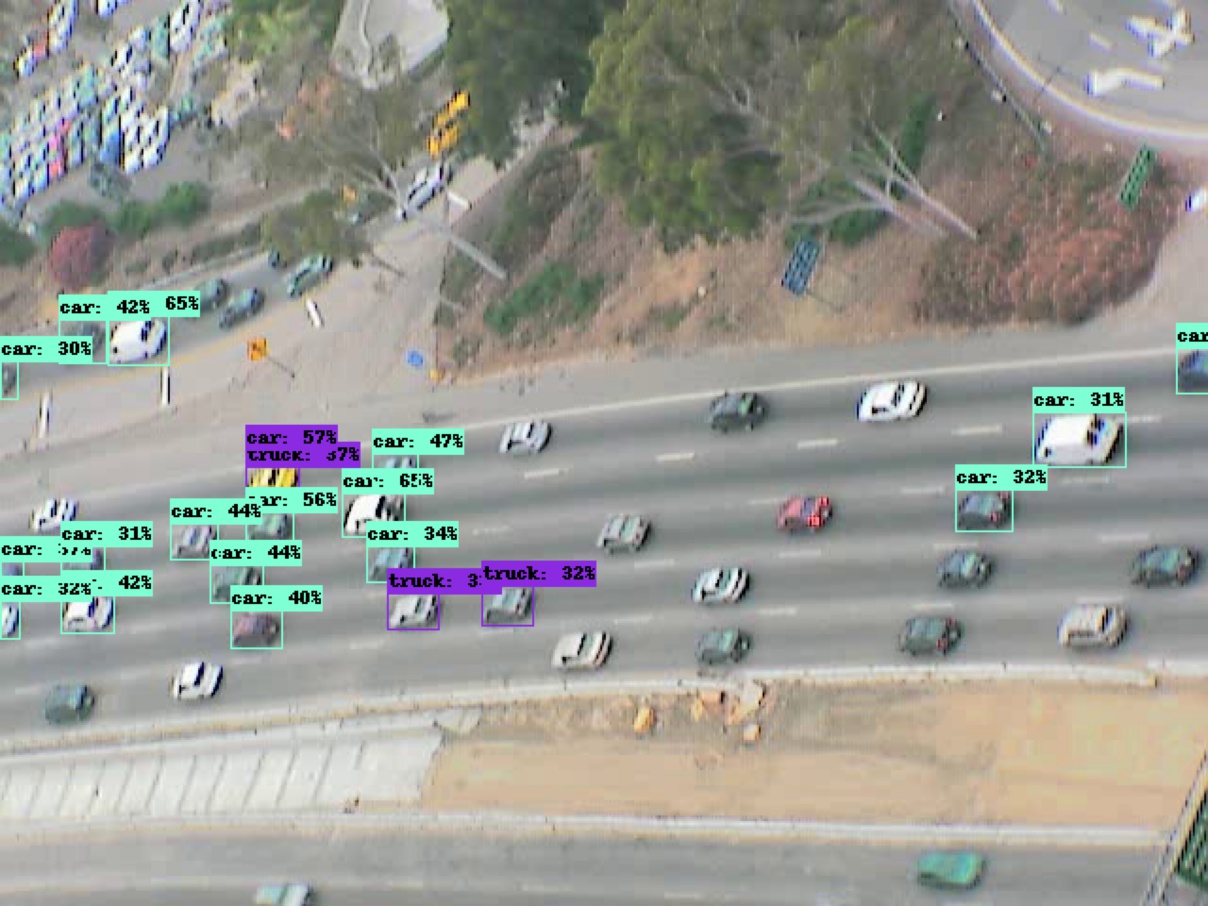}
   \end{minipage}\hfill
   \begin{minipage}{0.495\textwidth}
     \centering
     \includegraphics[width=1\linewidth]{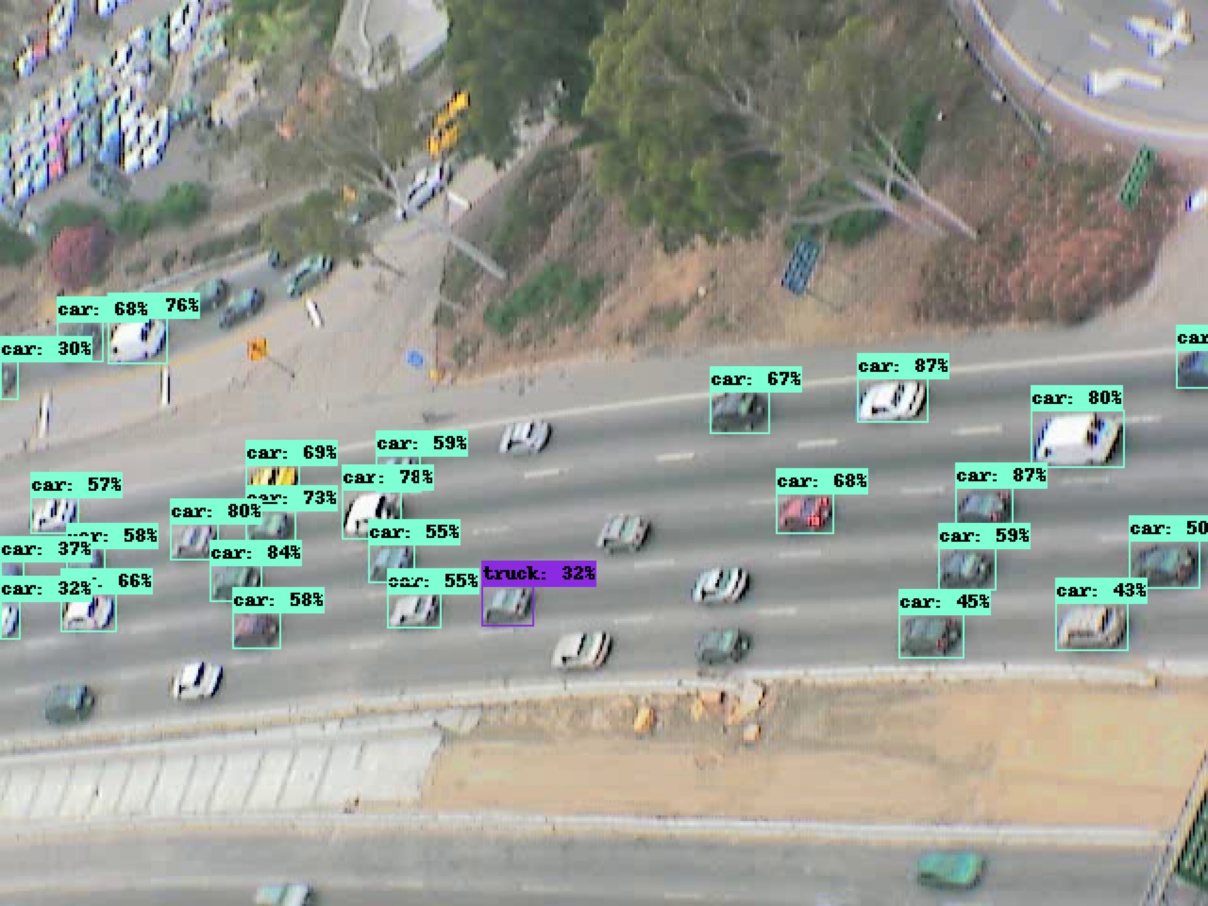}
   \end{minipage}
   \caption{Frame 1 of the first video denoted as sb-camera1-0820am-0835am. The left side shows the results obtained by the unmodified model while the right side shows the detections after applying our proposal.}
   \label{figure9}
\end{figure}

\begin{figure}[!ht]
   \begin{minipage}{0.495\textwidth}
     \centering
     \includegraphics[width=1\linewidth]{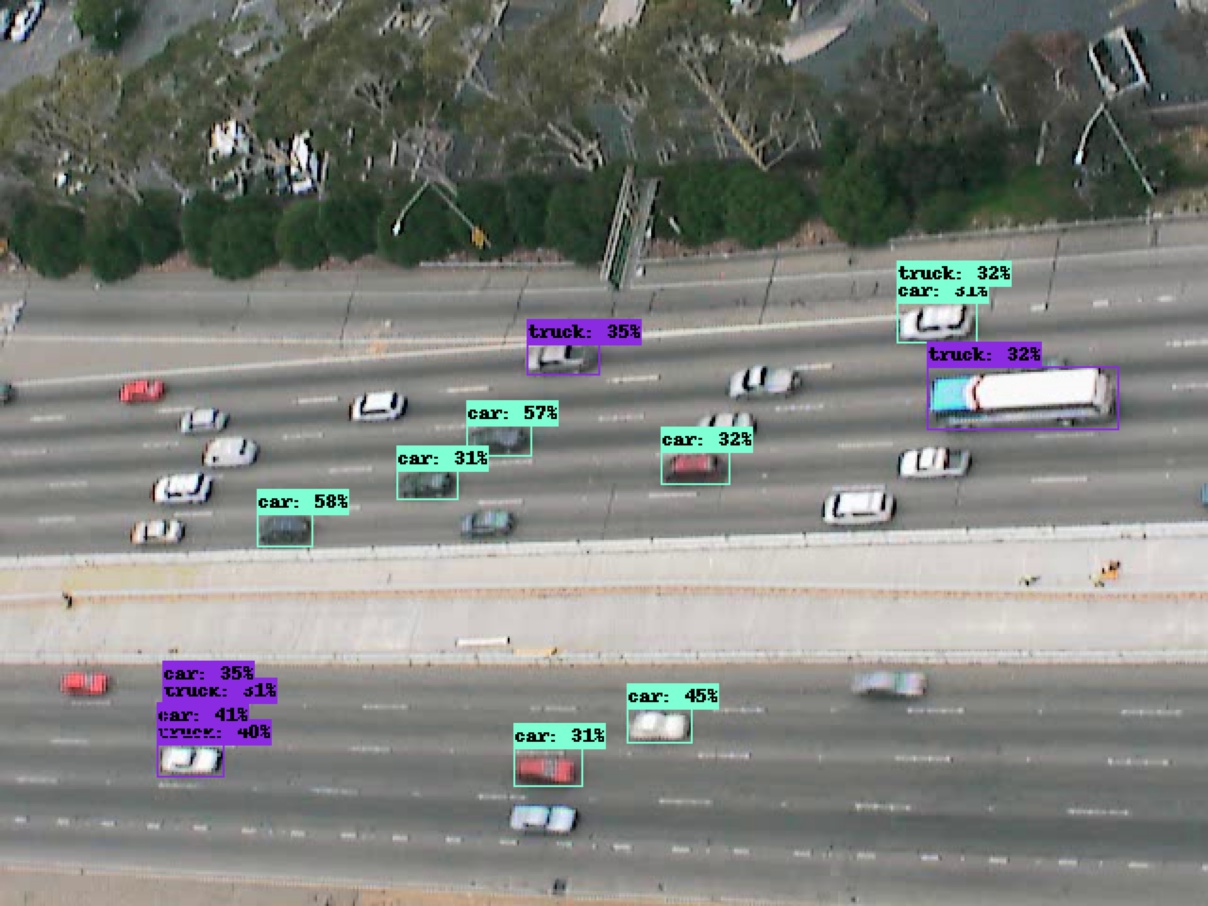}
   \end{minipage}\hfill
   \begin{minipage}{0.495\textwidth}
     \centering
     \includegraphics[width=1\linewidth]{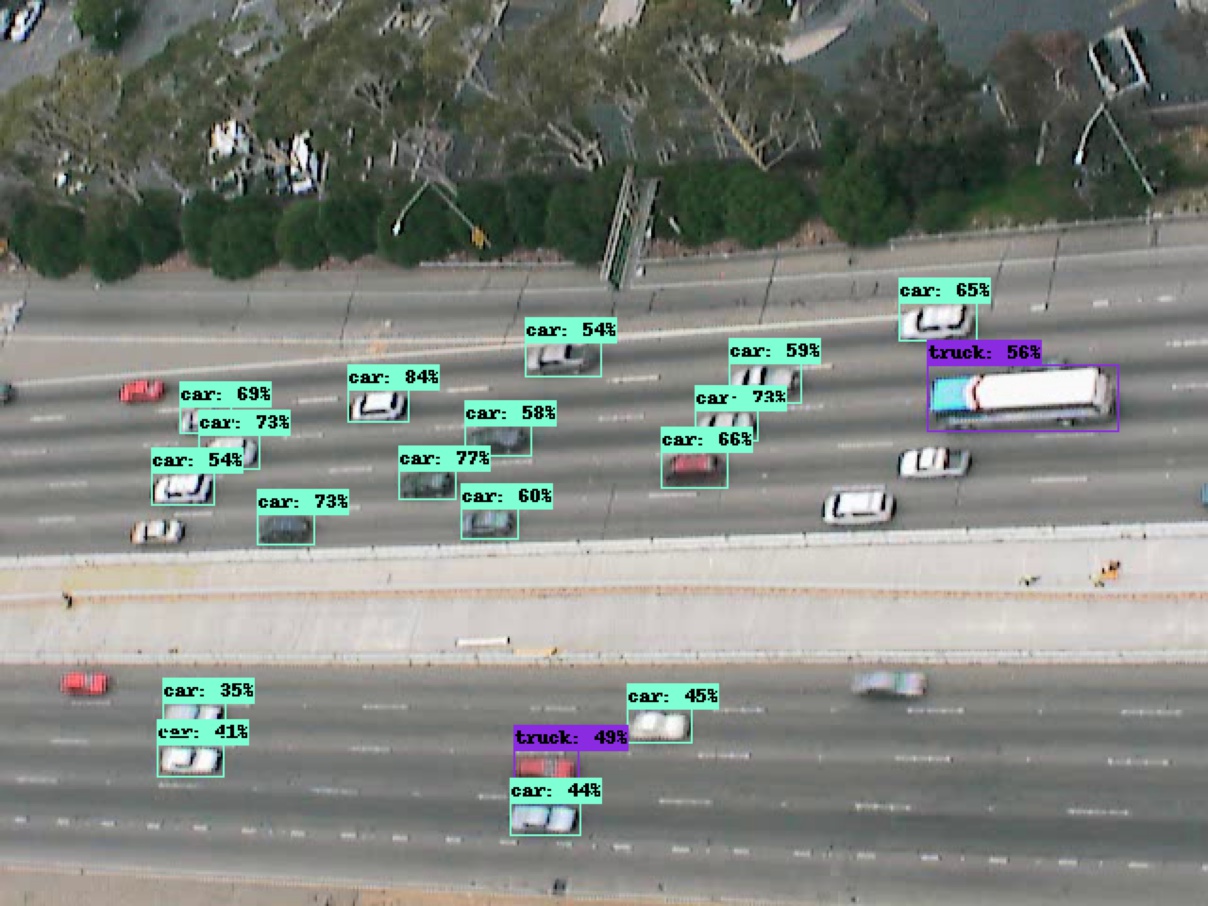}
   \end{minipage}
   \caption{Frame 6 of the second video denoted as sb-camera3-0820am-0835am. The left side shows the results obtained by the unmodified model while the right side shows the detections after applying our proposal using CenterNetHourGlass104Keypoints.}
    \label{figure10}
\end{figure}

\begin{figure}[!ht]
   \begin{minipage}{0.495\textwidth}
     \centering
     \includegraphics[width=1\linewidth]{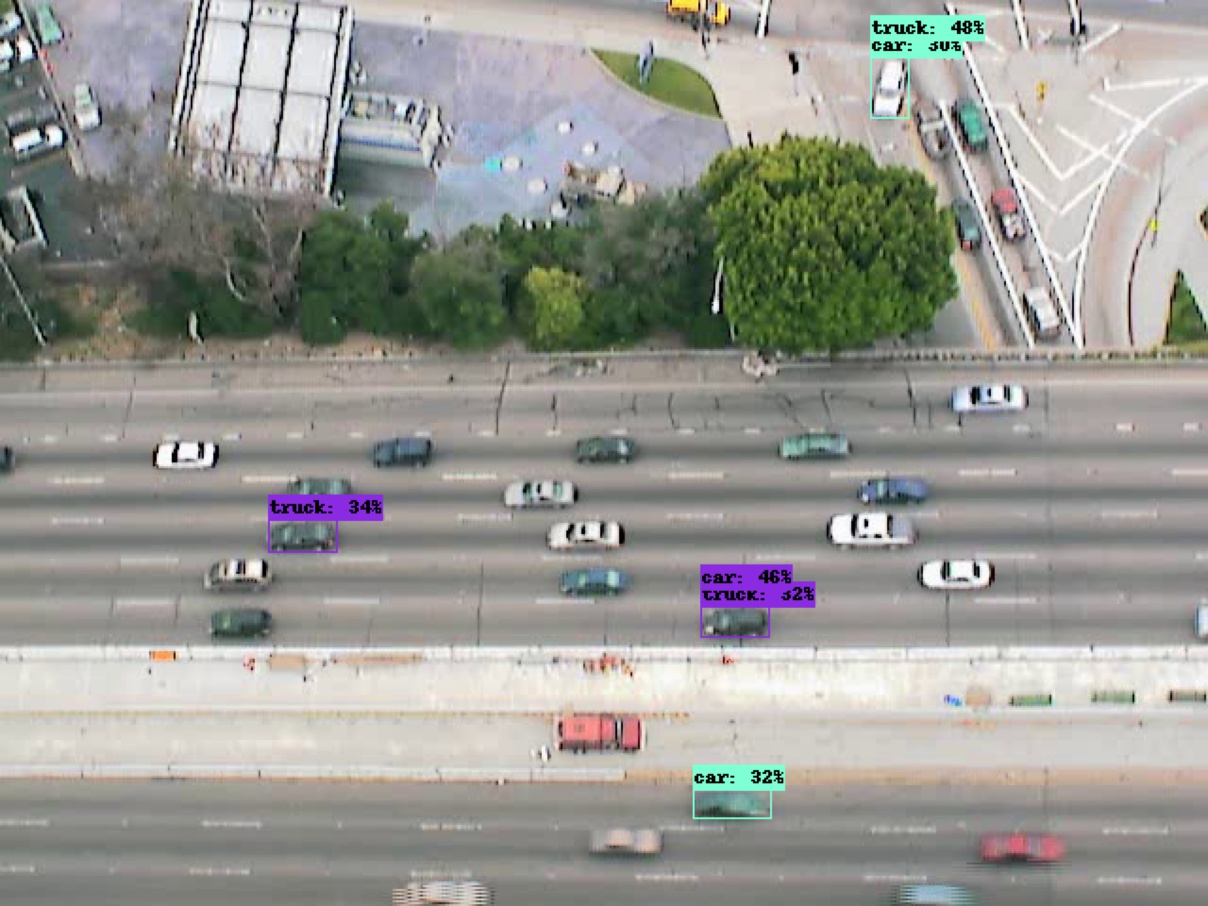}
   \end{minipage}\hfill
   \begin{minipage}{0.495\textwidth}
     \centering
     \includegraphics[width=1\linewidth]{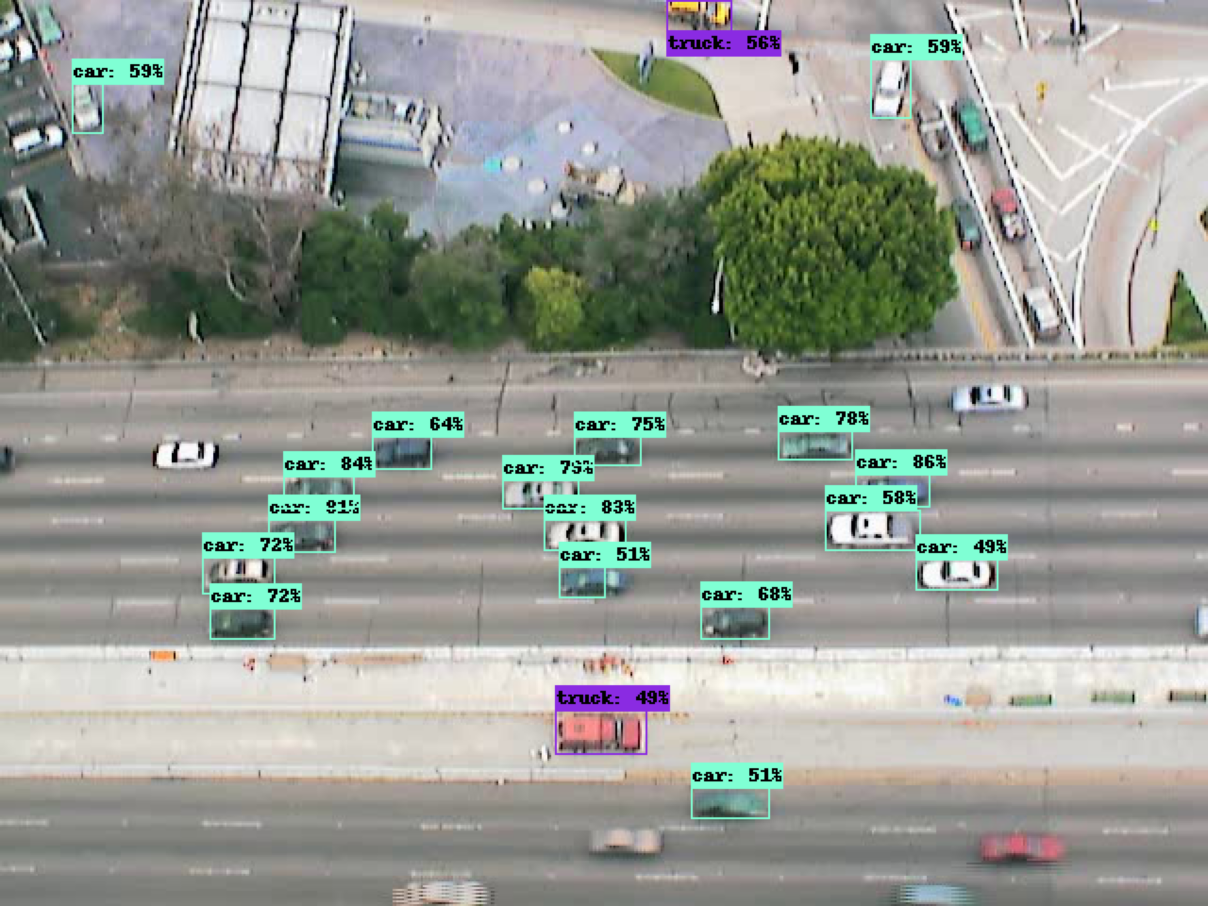}
   \end{minipage}
   \caption{Frame 136 of the third video denoted as sb-camera4-0820am-0835am. The left side shows the results obtained by the unmodified model while the right side shows the detections after applying our proposal using CenterNetHourGlass104Keypoints.}
    \label{figure11}
\end{figure}

\begin{figure}[!ht]
   \begin{minipage}{0.495\textwidth}
     \centering
     \includegraphics[width=1\linewidth]{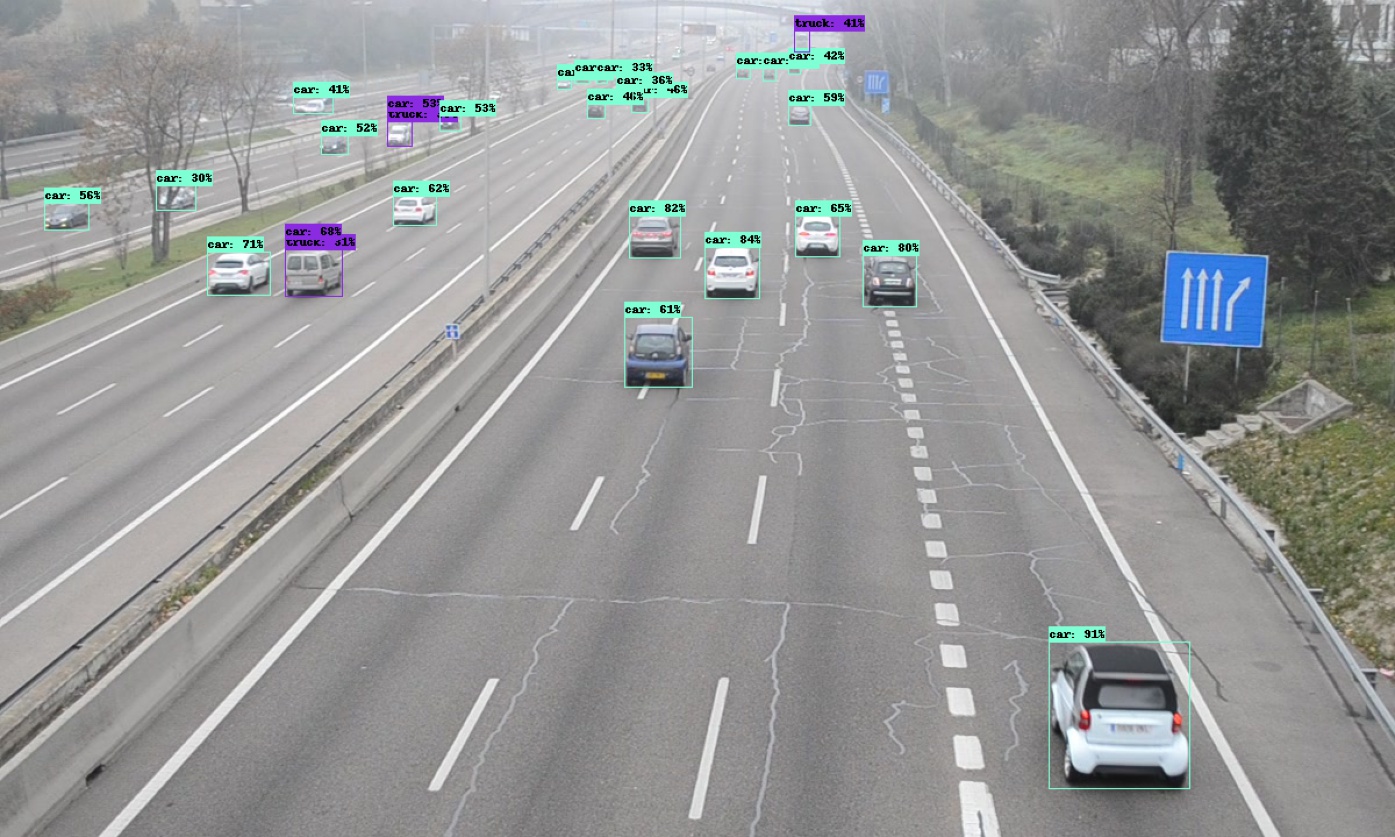}
   \end{minipage}\hfill
   \begin{minipage}{0.495\textwidth}
     \centering
     \includegraphics[width=1\linewidth]{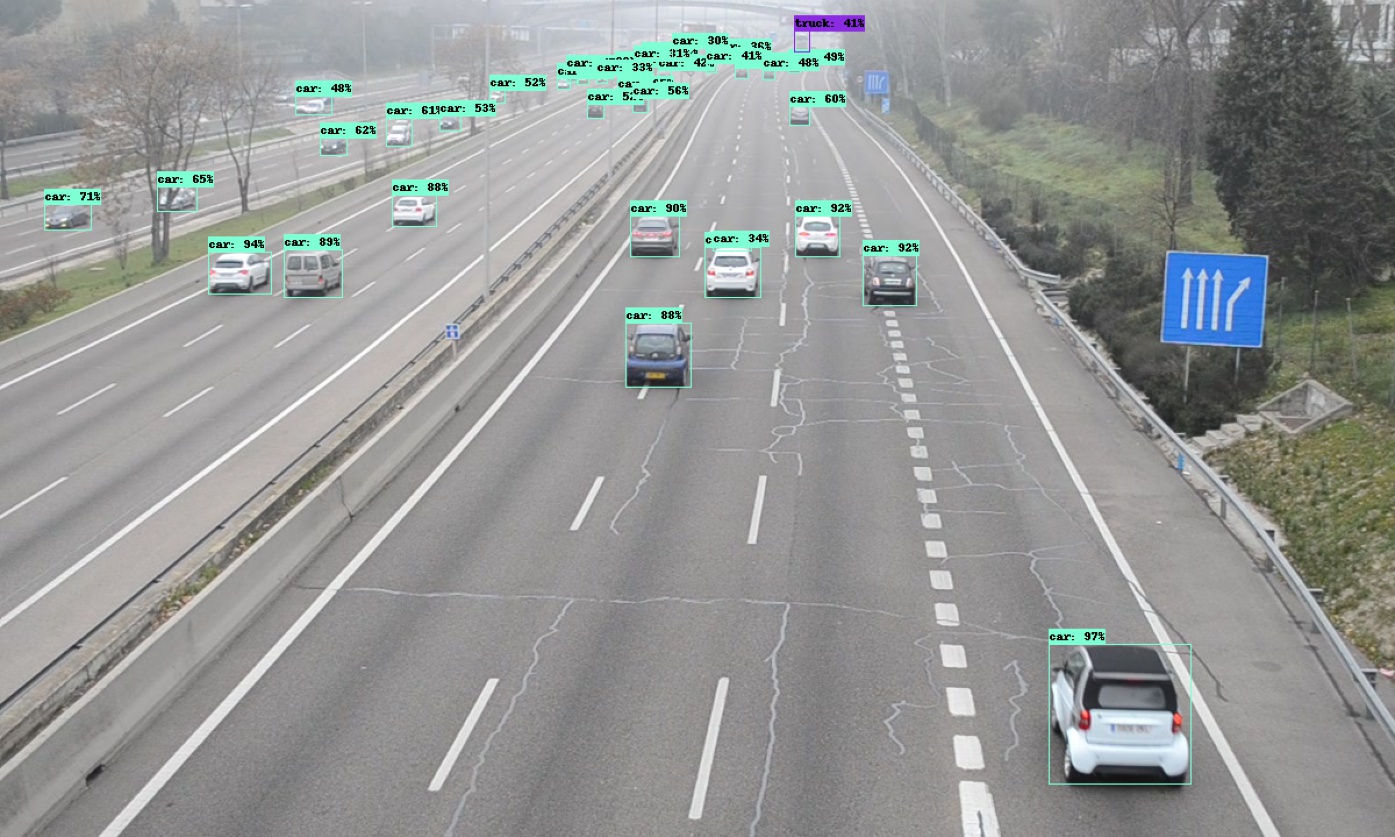}
   \end{minipage}
   \caption{Frame 5837 of the M-30-HD dataset \cite{M30HD}. The left side shows the results obtained by the unmodified model while the right side shows the detections after applying our proposal using CenterNetHourGlass104Keypoints.}
   \label{fig:M30}
\end{figure}

\begin{figure}[!ht]
  
   \begin{minipage}{0.495\textwidth}
     \centering
     \includegraphics[width=1\linewidth]{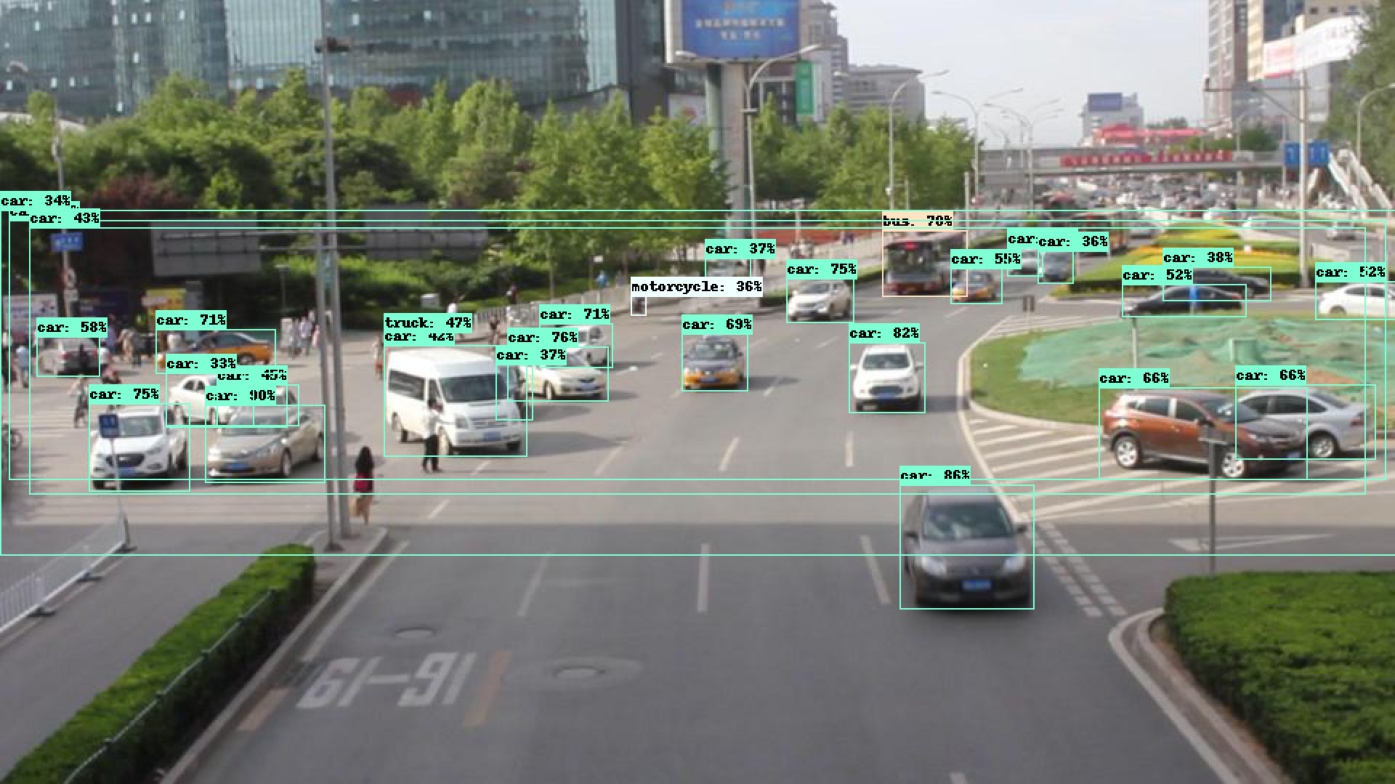}
   \end{minipage}\hfill
   \begin{minipage}{0.495\textwidth}
     \centering
     \includegraphics[width=1\linewidth]{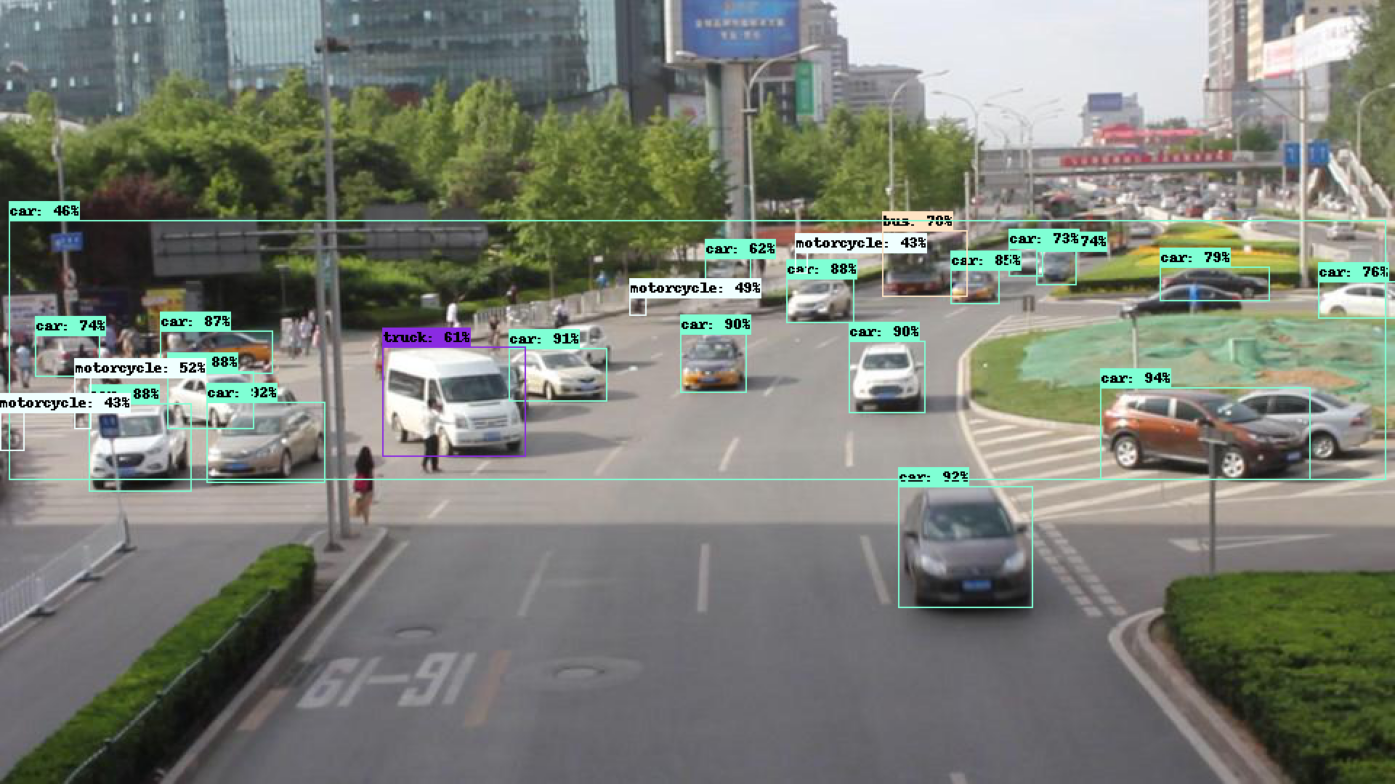}
   \end{minipage}
   \caption{Frame 2 of the UA-DETRAC MVI\_39311 dataset \cite{DETRAC1,DETRAC2,DETRAC3}. The left side shows the results obtained by the unmodified model while the right side shows the detections after applying our proposal using CenterNetHourGlass104Keypoints.}
   \label{fig:M2}
\end{figure}

\begin{figure}[!ht]
  
   \begin{minipage}{0.495\textwidth}
     \centering
     \includegraphics[width=1\linewidth]{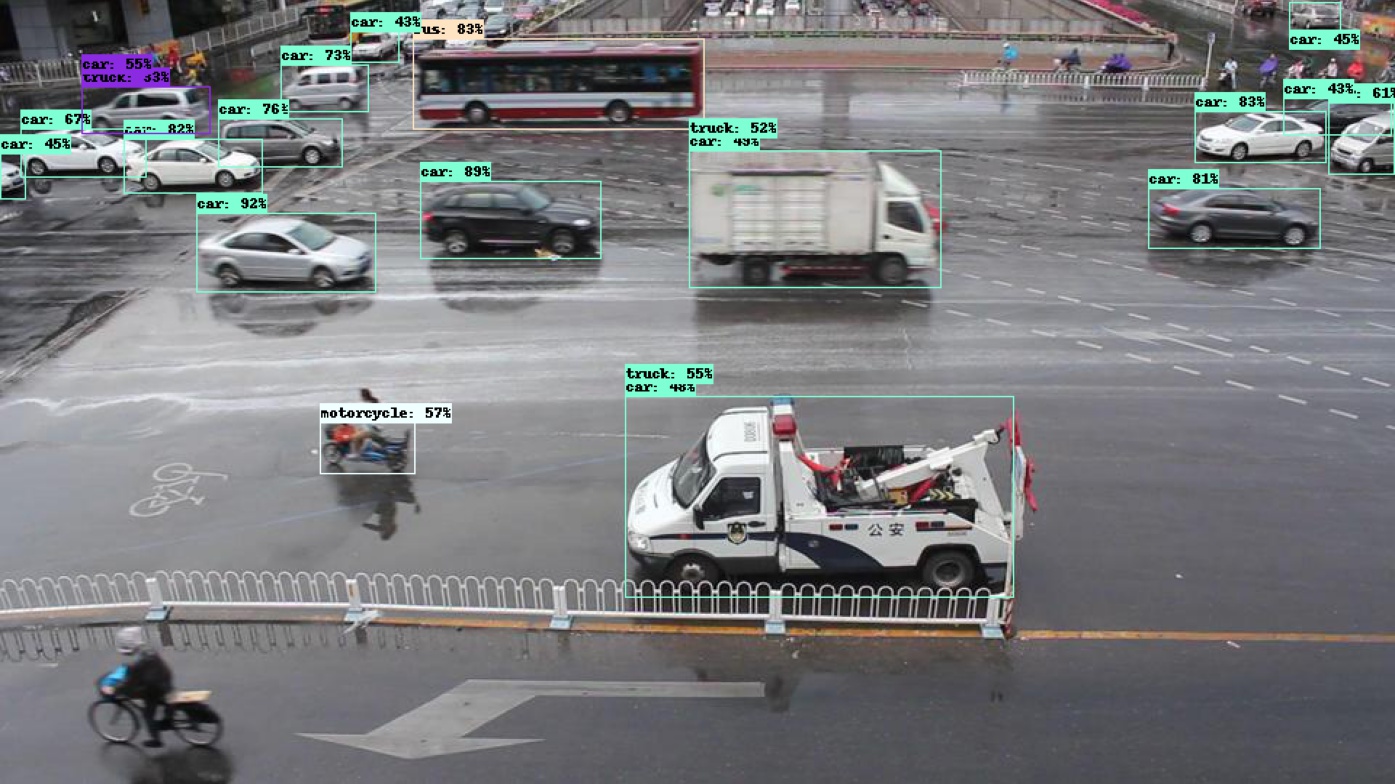}
   \end{minipage}\hfill
   \begin{minipage}{0.495\textwidth}
     \centering
     \includegraphics[width=1\linewidth]{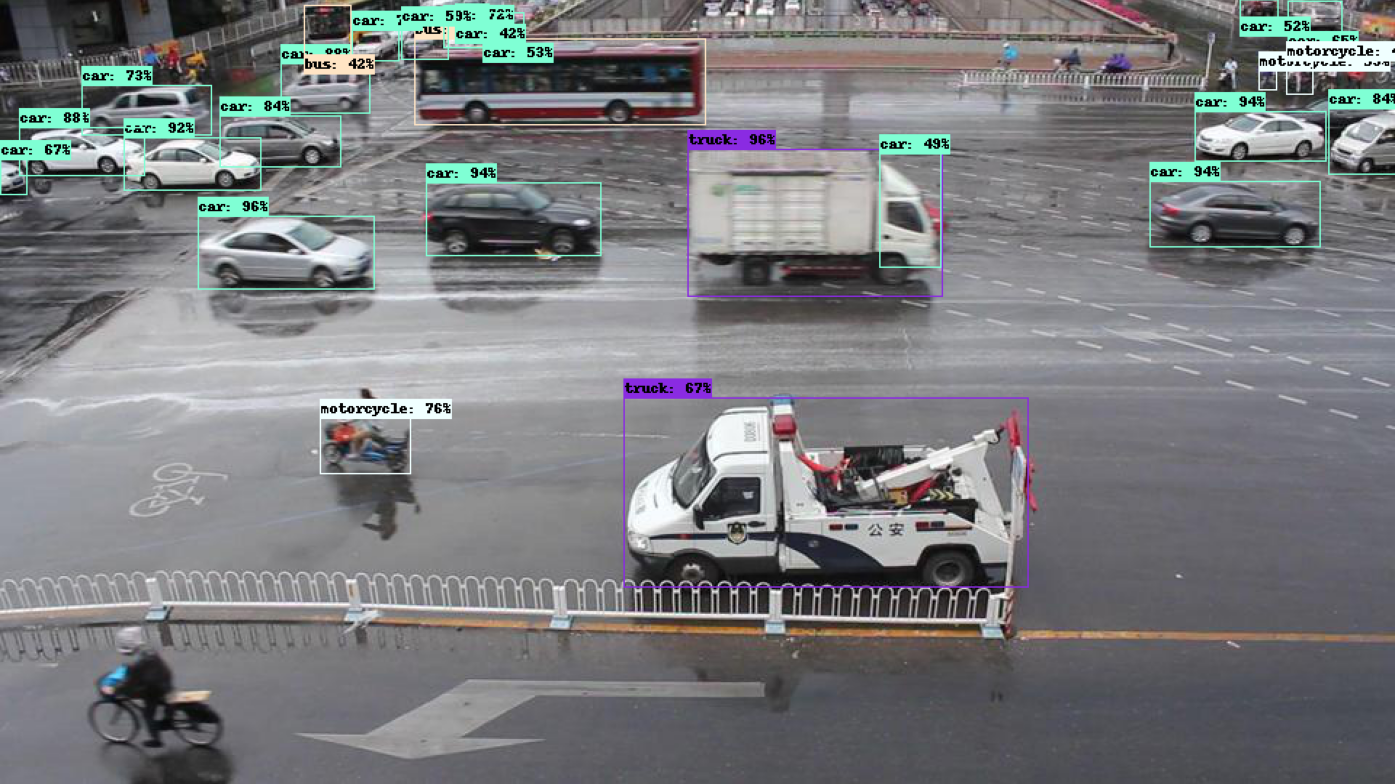}
   \end{minipage}
   \caption{Frame 1 of the UA-DETRAC MVI\_40903 dataset \cite{DETRAC1,DETRAC2,DETRAC3}. The left side shows the results obtained by the unmodified model while the right side shows the detections after applying our proposal using CenterNetHourGlass104Keypoints.}
   \label{fig:M22}
\end{figure}

Next, a series of examples are shown in Figures \ref{figure9},\ref{figure10},\ref{figure11},\ref{fig:M30}, \ref{fig:M2} and \ref{fig:M22} in which the qualitative results obtained with the proposed technique are shown for some instances of the considered dataset.

To check the results obtained by the proposed technique, in addition to the images coming to the selected videos, the technique has been tested on a series of previously available ground truth frames from other benchmark videos as can be seen in Figures \ref{fig:M30}, \ref{fig:M2} and \ref{fig:M22}. To obtain quantitative performance results, the evaluation process developed by \emph{COCO}\footnote{\url{https://github.com/cocodataset/cocoapi/blob/master/PythonAPI/pycocotools/cocoeval.py}} has been used. The standard measure for object detection tasks is the mean average precision (mAP), which takes into account the bounding box prediction and the class inference for each object. Thus, the bounding box of a prediction is considered a correct detection depending on the overlap with the ground truth bounding box, as measured by the intersection over the union known as \emph{IoU}. In addition, the class inference must be correct for that element. As output, several mAPs measures are obtained, divided by the object size and the quality of the inference class. According to the tests carried out, our technique solves a series of problems depending on the context of the image. Firstly, it detects many elements that were not detected a priori by the model in the first pass. Secondly, in the initial inference, we can observe in certain areas that the class of the elements is not correct, and after applying our technique, this problem is solved.

\begin{figure}[!htb]
   \begin{minipage}{0.495\textwidth}
     \centering
     \includegraphics[width=1\linewidth]{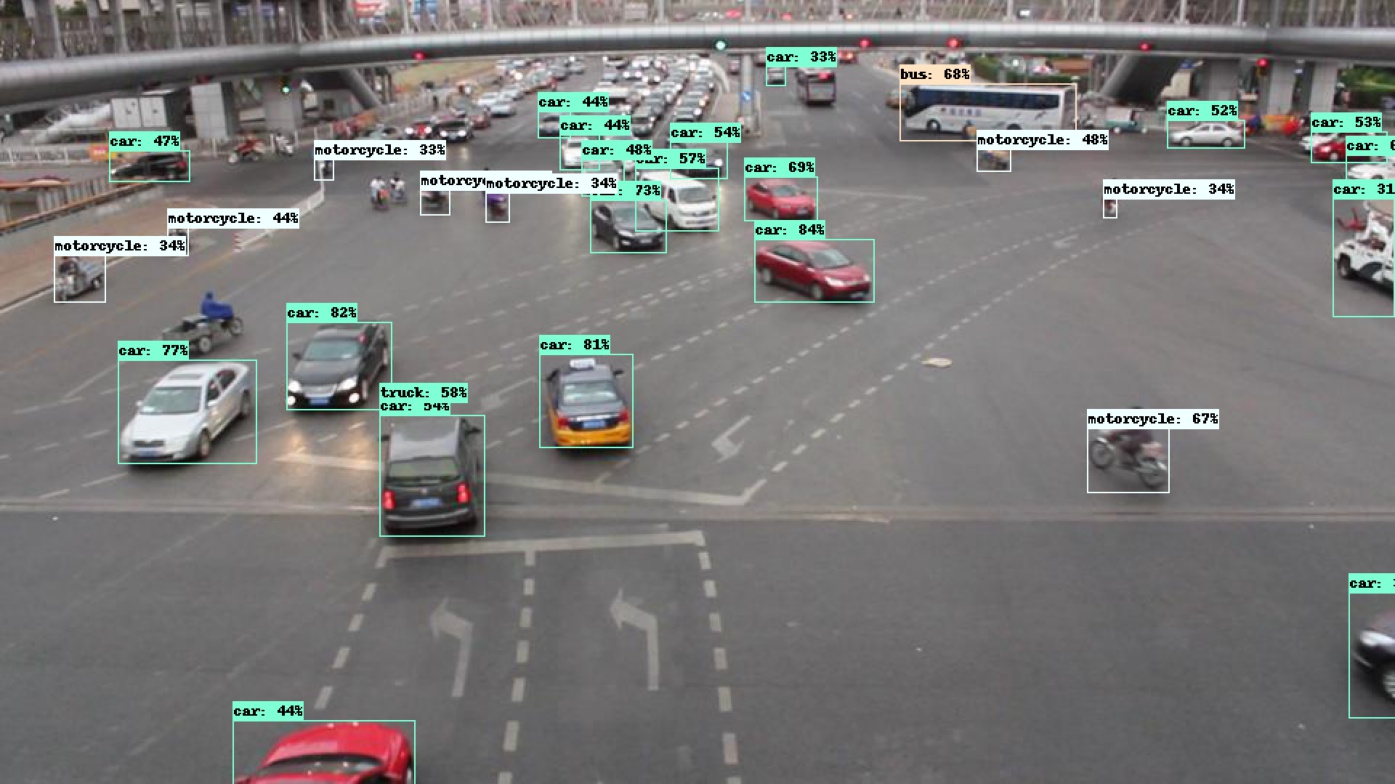}
   \end{minipage}\hfill
   \begin{minipage}{0.495\textwidth}
     \centering
     \includegraphics[width=1\linewidth]{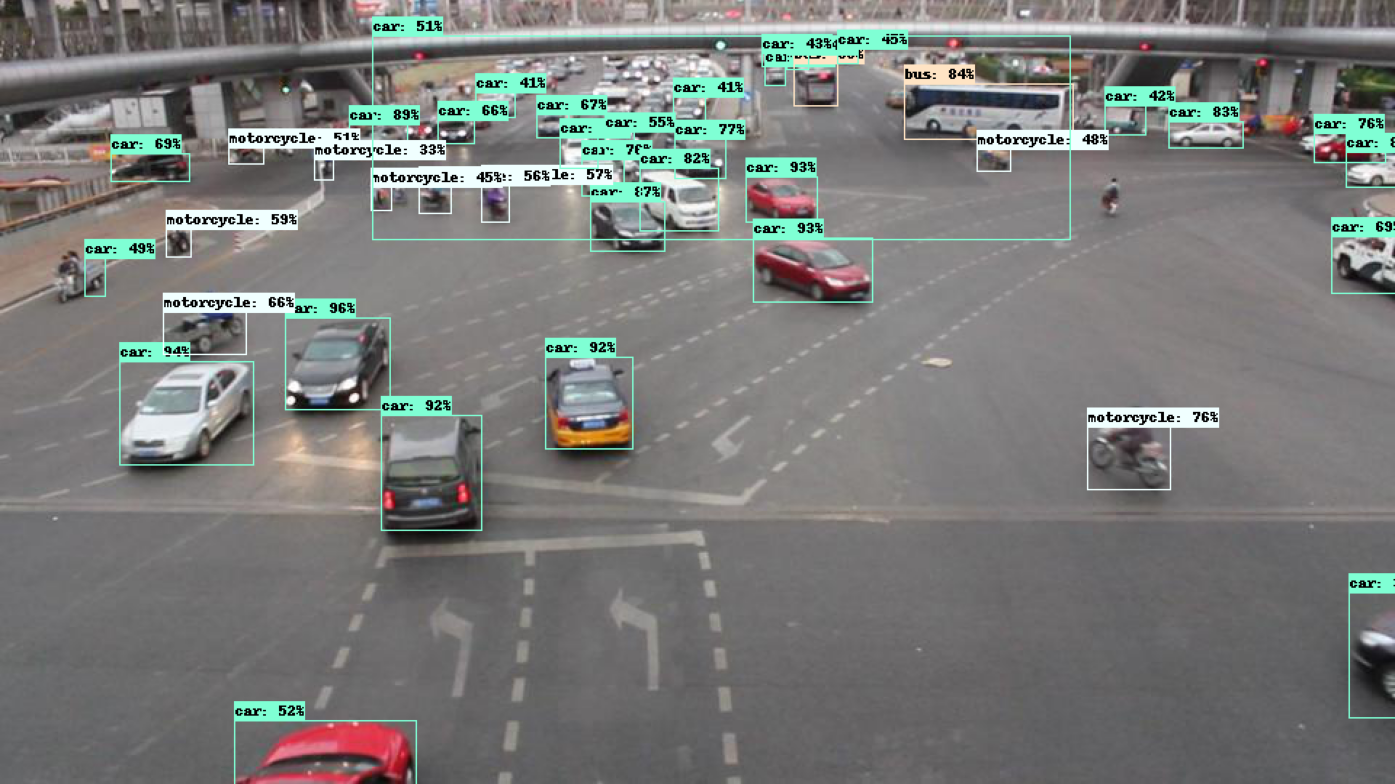}
   \end{minipage}
   \caption{Frame 5 of the UA-DETRAC MVI\_40852 dataset \cite{DETRAC1,DETRAC2,DETRAC3}. The left side shows the results obtained by the unmodified model while the right side shows the detections after applying our proposal using CenterNetHourGlass104Keypoints.}
   \label{Fig:SR10}
\end{figure}

 However, it should be noted that our solution is not infallible, and its performance depends on the context in which it is applied. For example, in Figure \ref{Fig:SR10} we can see how the number of elements is increased. However, we also obtain false positives such as the one in the center of the image or the incorrect inference of the motorcycle located at the center-left side. This problem can be alleviated by increasing the minimum percentage of inference to take into account a detection.

To carry out the tests, several variants of either the \emph{EfficientNet} or \emph{Centernet} models have been used. For \emph{EfficientNet}, some variants have been defined, which range from zero to seven, each representing variants for efficiency and accuracy according to a given scale. Thanks to this scaling heuristic, it is possible for the base model denoted as \emph{D0} to outperform the models at each scale, also avoiding an extensive search in the hyperparameter space. At first, one may get the impression that \emph{EfficientNet} constitutes a continuous family of models defined by an arbitrary choice of scale; however, other aspects such as resolution, width, and depth are contemplated. According to resolution, models with low scale such as \emph{D0} or \emph{D1} apply zero-padding near the boundaries of specific layers, thereby wasting computational resources. As for depth and width, the size of the channels must be a multiple of eight. For \emph{Centernet}, three variants have been used, each of them with a defined architecture, featuring as main differences the size of the image given as input as well as the way of processing each of the detections. It has been tested with variants based on keypoints as well as traditional models. These pre-trained models have been extracted from Tensorflow Model Zoo\footnote{\url{https://github.com/tensorflow/models/blob/master/research/object_detection/g3doc/tf2_detection_zoo.md}}. It is important to note that no training process using traffic sequences is performed on these pre-trained models.

The performance of these initial models and the proposed methodology is quantitatively compared. Since the \emph{car} class is majority,  the performance in the detection of cars is considered as the measure to compare. For this purpose, only class number 3 has been considered, corresponding in \emph{COCO} to the \emph{car} element, thus obtaining Tables \ref{tab:coches}, \ref{tab:coches1} and \ref{tab:coches2}. In each of these tables, the mean average precision (mAP) is determined for both the initial detections as well as the results obtained after applying our proposal. Columns second to fourth refer to detections of objects of any size. The second column corresponds to the mAP of the elements detected with an \emph{IoU} between 50 and 90 percent. The third column reports detections above 50 percent of \emph{IoU}, while the fourth column corresponds to the detections above 75 percent of this metric. Additionally, results are presented for small-sized objects with an \emph{IoU} between 50 and 95 percent and medium-sized items with an \emph{IoU} above 50 percent, in the fifth and sixth columns, respectively.

Metrics obtained for large-sized objects have been omitted since there are no samples in the established dataset. As shown in each of the tables obtained for the images that make up the defined dataset, there is a clear improvement in the mAP measure.

\begin{figure}[ht!]
\centering
\includegraphics[width=1\linewidth]{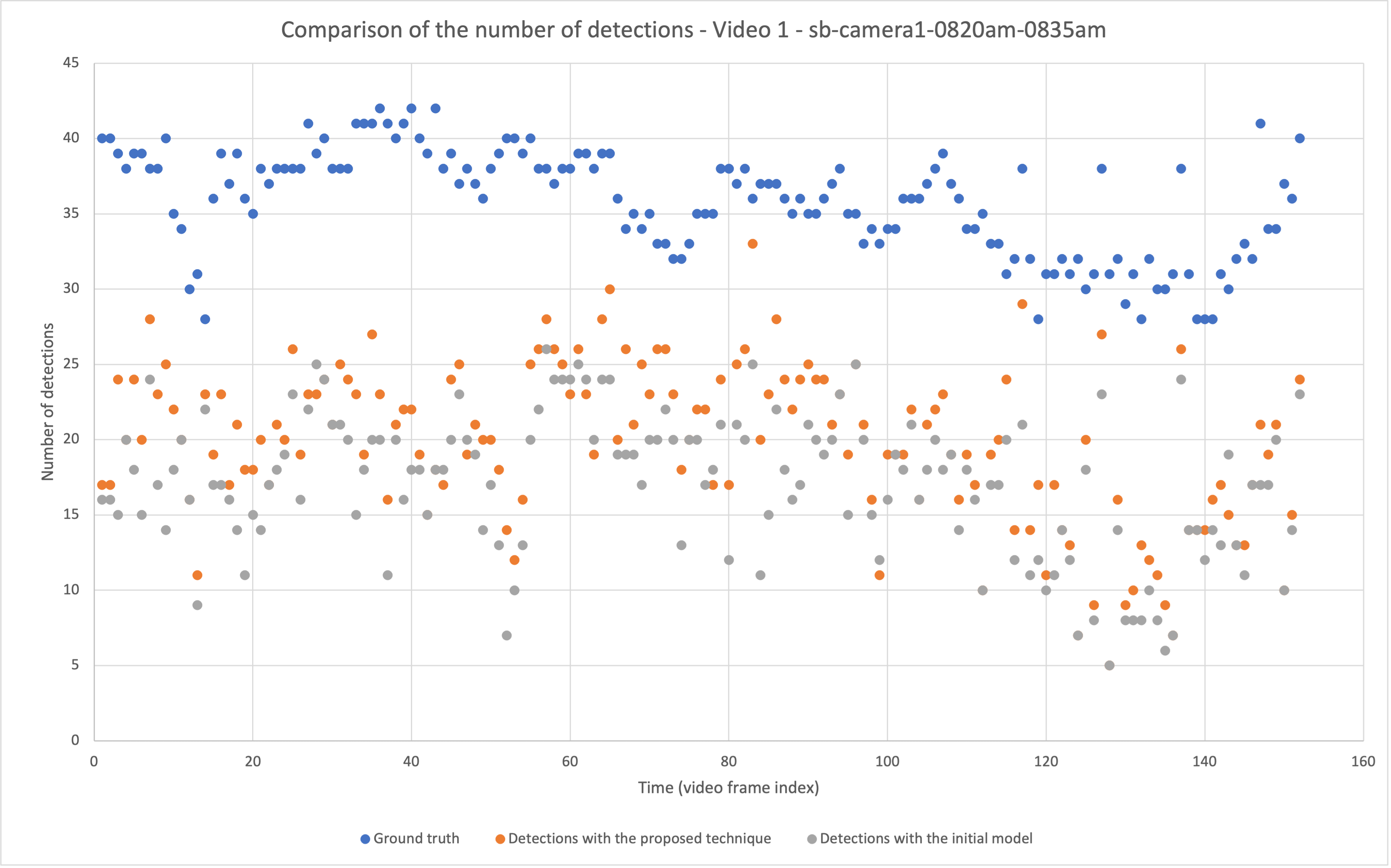}
\caption{Comparison of the number of detections for the first sequence.}
\label{fig:f11}
\end{figure}

\begin{figure}[ht!]
\centering
\includegraphics[width=1\linewidth]{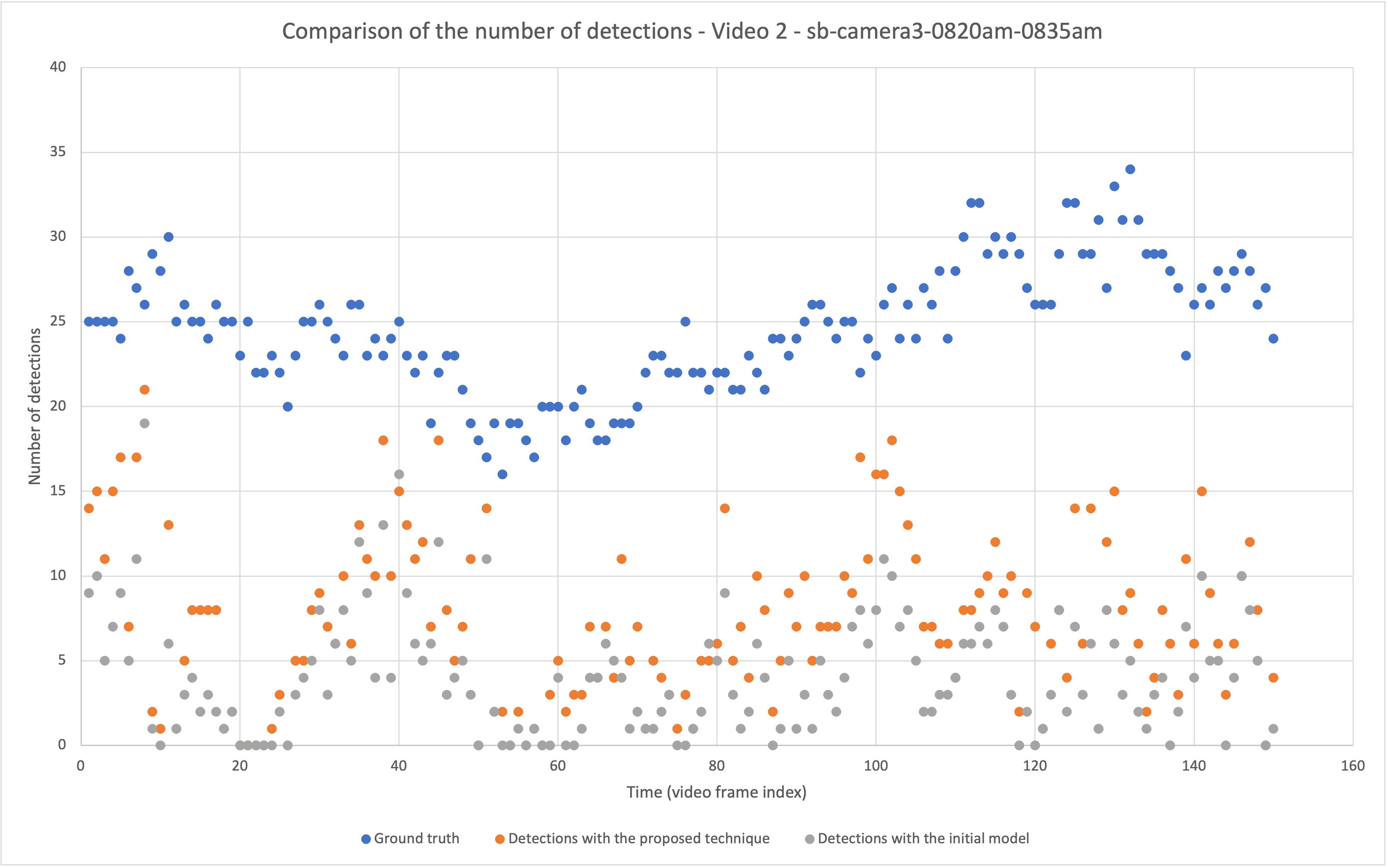}
\caption{Comparison of the number of detections for the second sequence.}
\label{fig:f12}
\end{figure}

\begin{figure}[ht!]
\centering
\includegraphics[width=1\linewidth]{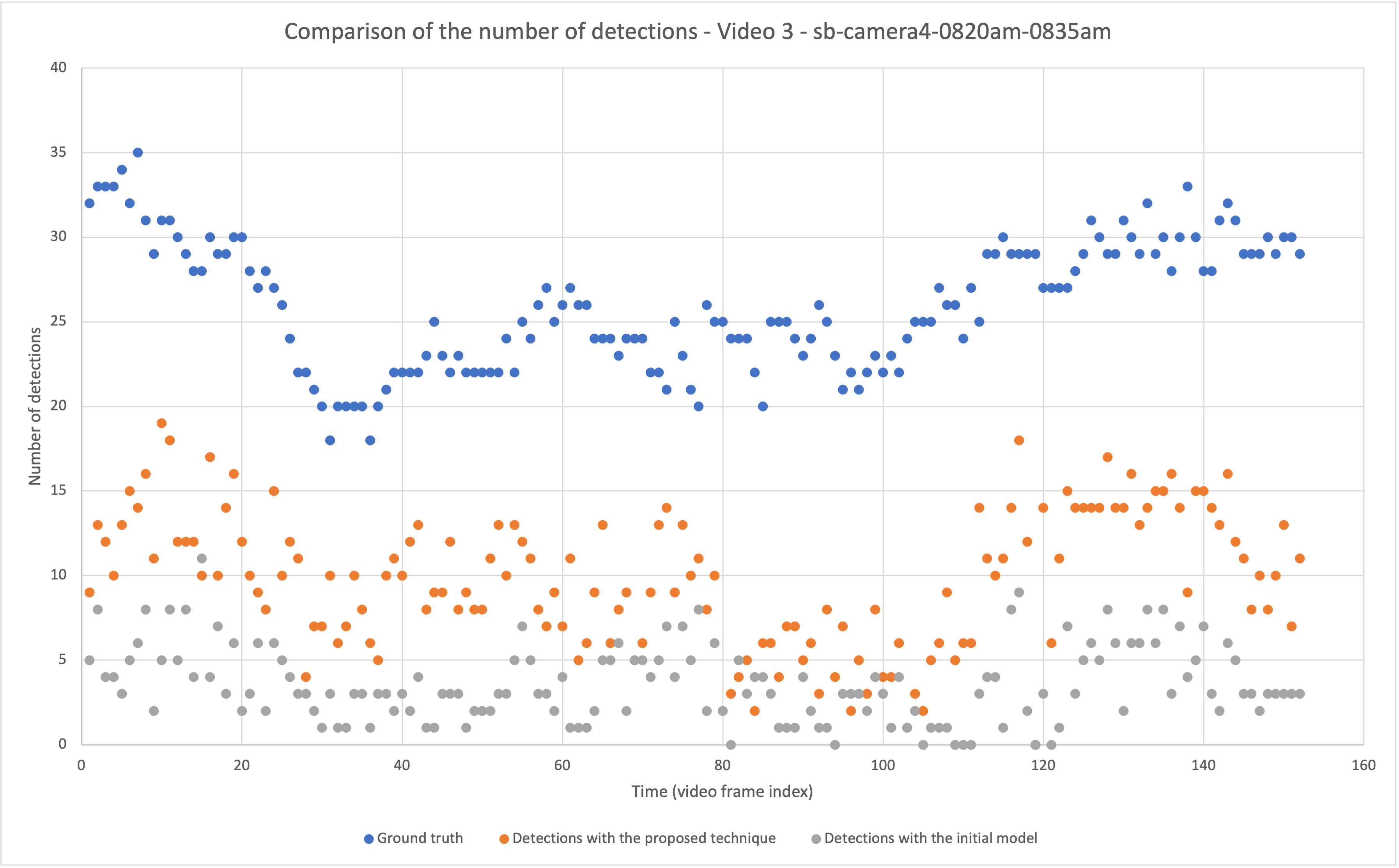}
\caption{Comparison of the number of detections for the third sequence.}
\label{fig:f13}
\end{figure}

To better illustrate the increase in the number of detections, the \emph{CenterNet HourGlass104 Keypoints} model has been selected for this test as it is one of the best performing models compared to the rest of the models used for this experiment. For each of the sets of images that make up the videos from which the dataset has been made, the improvement obtained in the number of elements by the described technique compared to the results initially provided by the model has been graphically represented in Figures \ref{fig:f11}, \ref{fig:f12} and \ref{fig:f13}.

\begin{sidewaystable}

\resizebox{19cm}{!} {
\begin{tabular}{| c | c | c | c | c | c | }
\hline
\multicolumn{6}{ |c| }{Video 1 - sb-camera1-0820am-0835am} \\ \hline
Model & IoU=0.50:0.95|area=all & IoU>0.50|area=all & IoU>0.75|area=all & IoU=0.50:0.95|area=Small & IoU>0.50|area=Medium  \\ \hline
CenterNet HourGlass104 Keypoints & 0.244 / \textbf{0.361} & 0.420 / \textbf{0.628} & 0.263 / \textbf{0.381} & 0.258 / \textbf{0.375} & 0.193 / \textbf{0.283}  \\
CenterNet MobileNetV2 FPN Keypoints & 0.083 / \textbf{0.161} & 0.191 / \textbf{0.358} & 0.053 / \textbf{0.121} & 0.085 / \textbf{0.163} & 0.081 / \textbf{0.122} \\
CenterNet Resnet101 V1 FPN & 0.273 / \textbf{0.350} & 0.512 / \textbf{0.656} & 0.248 / \textbf{0.318} & 0.274 / \textbf{0.355} & 0.244 / \textbf{0.273}  \\
EfficientDet D3 & 0.159 / \textbf{0.336} & 0.276 / \textbf{0.585} & 0.169 / \textbf{0.364} & 0.159 / \textbf{0.343} & 0.154 / \textbf{0.247}  \\
EfficientDet D4 & 0.236 / \textbf{0.442} & 0.424 / \textbf{0.790} & 0.237 / \textbf{0.459} & 0.243 / \textbf{0.451} & 0.168 / \textbf{0.298}  \\
EfficientDet D5 & 0.245 / \textbf{0.378} & 0.433 / \textbf{0.672} & 0.252 / \textbf{0.389} & 0.251 / \textbf{0.395} & 0.158 / \textbf{0.164}  \\ \hline
\end{tabular}
}
\caption{Results obtained for the first video. On the left is the mAP obtained by the base model and on the right of each box is the mAP obtained by our proposal (higher is better). The best results are marked in \textbf{bold}.}
\label{tab:coches}

\vspace{2\baselineskip}

\resizebox{19cm}{!} {
\begin{tabular}{| c | c | c | c | c | c | }
\hline
\multicolumn{6}{ |c| }{Video 2 - sb-camera3-0820am-0835am} \\ \hline
Model & IoU=0.50:0.95|area=all & IoU>0.50|area=all & IoU>0.75|area=all & IoU=0.50:0.95|area=Small & IoU>0.50|area=Medium  \\ \hline
CenterNet HourGlass104 Keypoints & 0.090 / \textbf{0.186} & 0.176 / \textbf{0.363} & 0.074 / \textbf{0.154} & 0.092 / \textbf{0.191} & 0.043 / \textbf{0.112}  \\
CenterNet MobileNetV2 FPN Keypoints & 0.117 / \textbf{0.158} & 0.309 / \textbf{0.390} & 0.047 / \textbf{0.083} & 0.119 / \textbf{0.159} & 0.039 / \textbf{0.085} \\
CenterNet Resnet101 V1 FPN & 0.111 / \textbf{0.188} & 0.242 / \textbf{0.403} & 0.072 / \textbf{0.124} & 0.111 / \textbf{0.191} & \textbf{0.096} / 0.093   \\
EfficientDet D3 & 0.092 / \textbf{0.294} & 0.177 / \textbf{0.590} & 0.073 / \textbf{0.224} & 0.091 / \textbf{0.297} & 0.105 / \textbf{0.171}  \\
EfficientDet D4 & 0.236 / \textbf{0.345} & 0.480 / \textbf{0.689} & 0.171 / \textbf{0.261} & 0.237 / \textbf{0.350} & 0.232 / \textbf{0.244}\\
EfficientDet D5 & 0.178 / \textbf{0.292} & 0.345 / \textbf{0.588} & 0.145 / \textbf{0.219} & 0.178 / \textbf{0.295} & 0.207 / \textbf{0.221} \\ \hline
\end{tabular}
}
\caption{Results obtained for the second video. On the left is the mAP obtained by the base model and on the right of each box is the mAP obtained by our proposal (higher is better). The best results are marked in \textbf{bold}.}
\label{tab:coches1}

\vspace{2\baselineskip}

\resizebox{19cm}{!} {
\begin{tabular}{| c | c | c | c | c | c | }
\hline
\multicolumn{6}{ |c| }{Video 3 - sb-camera4-0820am-0835am} \\ \hline
Model & IoU=0.50:0.95|area=all & IoU>0.50|area=all & IoU>0.75|area=all & IoU=0.50:0.95|area=Small & IoU>0.50|area=Medium  \\ \hline
CenterNet HourGlass104 Keypoints & 0.074 / \textbf{0.219} & 0.137 / \textbf{0.394} & 0.072 / \textbf{0.217} & 0.075 / \textbf{0.227} & 0.054 / \textbf{0.060} \\
CenterNet MobileNetV2 FPN Keypoints & 0.060 / \textbf{0.102} & 0.159 / \textbf{0.237} & 0.026 / \textbf{0.064} & 0.066 / \textbf{0.108} & 0.009 / \textbf{0.014} \\
CenterNet Resnet101 V1 FPN & 0.039 / \textbf{0.082} & 0.075 / \textbf{0.160} & 0.028 / \textbf{0.065} & 0.040 / \textbf{0.084} & \textbf{0.016} / 0.015 \\
EfficientDet D3 & 0.029 / \textbf{0.163} & 0.050 / \textbf{0.292} & 0.022 / \textbf{0.164} & 0.029 / \textbf{0.168} & 0.012 / \textbf{0.040} \\
EfficientDet D4 & 0.129 / \textbf{0.280} & 0.236 / \textbf{0.518} & 0.118 / \textbf{0.264} & 0.130 / \textbf{0.287} & 0.068 / \textbf{0.151}\\
EfficientDet D5 & 0.056 / \textbf{0.204} & 0.099 / \textbf{0.384} & 0.051 / \textbf{0.184} & 0.058 / \textbf{0.207} & 0.006 / \textbf{0.082} \\ \hline
\end{tabular}
}
\caption{Results obtained for the third video. On the left is the mAP obtained by the base model and on the right of each box is the mAP obtained by performing our proposal (higher is better). The best results are marked in \textbf{bold}.}
\label{tab:coches2}
\end{sidewaystable}

\clearpage

According to Tables \ref{tab:coches}, \ref{tab:coches1}, \ref{tab:coches2} , and Figures \ref{fig:f11}, \ref{fig:f12} and \ref{fig:f13}, it can be stated that the application of the technique described in this work not only improves the accuracy of the elements initially detected by the model, but also detects objects that were not identified a priori. Thus, we can corroborate that in each and every one of the frames that make up the test videos, we have obtained a larger number of elements, denoted in the graphs by the orange dots, compared to the initial detections established by the gray dots.

\section{Conclusions}\label{conclusion}

Throughout this paper, a new technique has been developed which employs deep convolutional neural networks for super-resolution to detect small-scale visual elements and improve the reliability of the estimation of their classes. The method makes a first pass of the object detection deep neural network. Then it super-resolves the video frame. After that, a window is defined for each detected object, which has the detected object at its center, and a object detection is carried out for that window. This object detection results in more accurate estimations of the characteristics of the previously detected objects, as well as the detection of additional objects that were missed by the object detection network in the first pass.

A quantitative comparison of our approach has been carried out. The results obtained show that the application of super-resolution processes does improve the detections made on an image since, based on the metrics obtained using the \emph{COCO} evaluator, the quantitative object detection results are enhanced. Qualitative results further confirm the advantages of our proposal since the improvement of the reliability of the object detections and the discovery of additional objects can be observed.

Let us remember that, in all cases, the images that make up the model on which the results have been validated correspond to video surveillance systems placed in high points. Regarding future lines to extend the research reported in this article, it should be noted that a novel approach is envisaged on which work is already underway. It consists of reducing the number of windows that need to be processed during the object detection. This enhancement might result in a reduction of the computational load associated with the execution of our method.

\printbibliography

\nocite{*}

\vspace{3em}

\end{document}